%% file: main.tex
\documentclass{article}

\usepackage{microtype}
\usepackage{graphicx}
\usepackage{subcaption}
\usepackage{booktabs} 
\usepackage{tabularx}

\usepackage{hyperref}

\usepackage[preprint]{icml2026}

\usepackage{amsmath}
\usepackage{amssymb}
\usepackage{mathtools}
\usepackage{amsthm}

\usepackage[capitalize,noabbrev]{cleveref}

\theoremstyle{plain}

\theoremstyle{definition}

\theoremstyle{remark}

\usepackage[textsize=tiny]{todonotes}

\usepackage{xcolor}
\usepackage{amssymb}
\usepackage{arydshln}
\newcommand{\name}{\textbf{MultiBO}}
\newcommand{\checkmarkgreen}{\textcolor{green!60!black}{\checkmark}}
\newcommand{\crossred}{\textcolor{red}{\ding{55}}}
\usepackage{pifont}
\usepackage{wrapfig}
\usepackage{makecell}
\usepackage{amssymb}
\usepackage{comment}

\input{math_commands}

\usepackage{hyperref}
\usepackage{url}
\usepackage{booktabs} 
\usepackage{graphicx}
\usepackage{adjustbox}
\usepackage{multirow}
\usepackage{tcolorbox}
\usepackage{amsthm}
\usepackage{enumitem}
\usepackage{soul}
\usepackage[ruled,vlined,algo2e]{algorithm2e}
\usepackage{tcolorbox}
\usepackage{wrapfig}

\newcommand{\redbox}[1]{%
  \begin{tcolorbox}[colback=red!15,colframe=red!70!black,boxrule=0.8pt,arc=4pt,left=2pt,right=2pt,top=2pt,bottom=2pt]
  #1
  \end{tcolorbox}
}

\begin{document}

\twocolumn[
  \icmltitle{Personalized Image Generation via Human-in-the-loop Bayesian Optimization}


  \icmlsetsymbol{equal}{*}

  \begin{icmlauthorlist}
    \icmlauthor{Rajalaxmi Rajagopalan}{1}
    \icmlauthor{Debottam Dutta}{1}
    \icmlauthor{Yu-Lin Wei}{1}
    \icmlauthor{Romit Roy Choudhury}{1}
  \end{icmlauthorlist}

  \icmlaffiliation{1}{University of Illinois, Urbana-Champaign}

  \icmlcorrespondingauthor{Rajalaxmi Rajagopalan}{rr30@illinois.edu}

  \icmlkeywords{Machine Learning, ICML}

  \vskip 0.3in
]



\printAffiliationsAndNotice{}  


\newcommand*\circled[1]{\tikz[baseline=(char.base)]{%
    \node[shape=circle,fill=black,draw,inner sep=1pt,text=white] (char) {#1};}}


\input{tex/_0_romit_abstract}

\input{tex/_2_background}
\input{tex/_3_method}

\input{tex/_4_results}

\input{tex/_5_related_works}
\input{tex/conclusion}
\bibliography{reference}
\bibliographystyle{icml2026}

\clearpage
\appendix
\input{tex/appendix}

\end{document}

%% file: math_commands.tex
\usepackage{amsmath,amsfonts,bm}

\def\eqref#1{equation~\ref{#1}}

\def\1{\bm{1}}

\DeclareMathAlphabet{\mathsfit}{\encodingdefault}{\sfdefault}{m}{sl}
\SetMathAlphabet{\mathsfit}{bold}{\encodingdefault}{\sfdefault}{bx}{n}

\DeclareMathOperator*{\argmax}{arg\,max}

\newcommand{\greenbox}[1]{%
    \begin{tcolorbox}[colback=green!15, colframe=gray!30, arc=0pt, boxrule=0.5pt, left=-3pt, right=0pt, top=1pt, bottom=1pt]
    #1
    \end{tcolorbox}
}

%% file: tex/_0_romit_abstract.tex
\begin{abstract}
Imagine Alice has a specific image $x^*$ in her mind, say, the view of the street in which she grew up during her childhood.
To generate that exact image, she guides a generative model with multiple rounds of prompting and arrives at an image $x^{p*}$.
Although $x^{p*}$ is reasonably close to $x^*$, Alice finds it difficult to close that gap using language prompts.
This paper aims to narrow this gap by observing that even after language has reached its limits, humans can still tell when a new image $x^+$ is closer to $x^*$ than $x^{p*}$. 
Leveraging this observation, we develop {\name} (Multi-Choice Preferential Bayesian Optimization) that carefully generates $K$ new images as a function of $x^{p*}$, gets preferential feedback from the user, uses the feedback to guide the diffusion model, and ultimately generates a new set of $K$ images.
We show that within $B$ rounds of user feedback, it is possible to arrive much closer to $x^*$, even though the generative model has no information about $x^*$.
Qualitative scores from $30$ users, combined with quantitative metrics compared across $5$ baselines, show promising results, suggesting that multi-choice feedback from humans can be effectively harnessed for personalized image generation.
\vspace{-0.15in}
\end{abstract}

\section{Introduction}
Modern diffusion models like FLUX \cite{labs2025flux1kontextflowmatching}, StableDiffusion3 \cite{esser2024scaling}, Lumiere \cite{bartal2024lumierespacetimediffusionmodel} continue to make remarkable advances in image generation.
In tandem, users continue to raise the bar, asking for images to not only be of high quality, but to also accurately obey their language prompts;
the expectation is that the generated image will match an image $x^*$ that the user has in her mind.
Can generative models deliver on this expectation?
Perhaps detailed surgery on generative model outputs, using a careful combination of prompting, masking, editing, inpainting, etc. could produce the desired image.
However, for most lay users, language based prompting is the main form of expression and that may be inadequate for hyper-personalization.
This is expected because the space of images is much higher dimensional than language, hence a given prompt maps to many possible images. 
Moreover, humans have limited vocabulary and expressivity in describing an image, so not everyone will be able to perfectly craft the optimal prompt. 
Finally, in the near term, models that jointly learn language and image representations (e.g., CLIP) are likely to have imperfections, causing further misalignment between language and images.
Assuming these are true, it appears that there will be a fundamental gap between the image $x^*$ the user has in mind, and the best image $x^{p*}$ that the user can generate through language prompts $p$. 
Can we narrow down this gap for everyday users? 

Observe that even after language-based prompting has saturated, if a diffusion model could present some images better than $x^{p*}$, humans can quite robustly pick the image closest to their target $x^*$. 
This preference indication carries valuable information to narrow the gap. 
We cast this as a human-in-the-loop black box optimization problem with preferential input. 
The goal is to achieve free-form, training-free and interactive image personalization, specifically when language prompting has neared saturation.

\begin{figure*}[t]
    \centering
    \includegraphics[width=\linewidth]{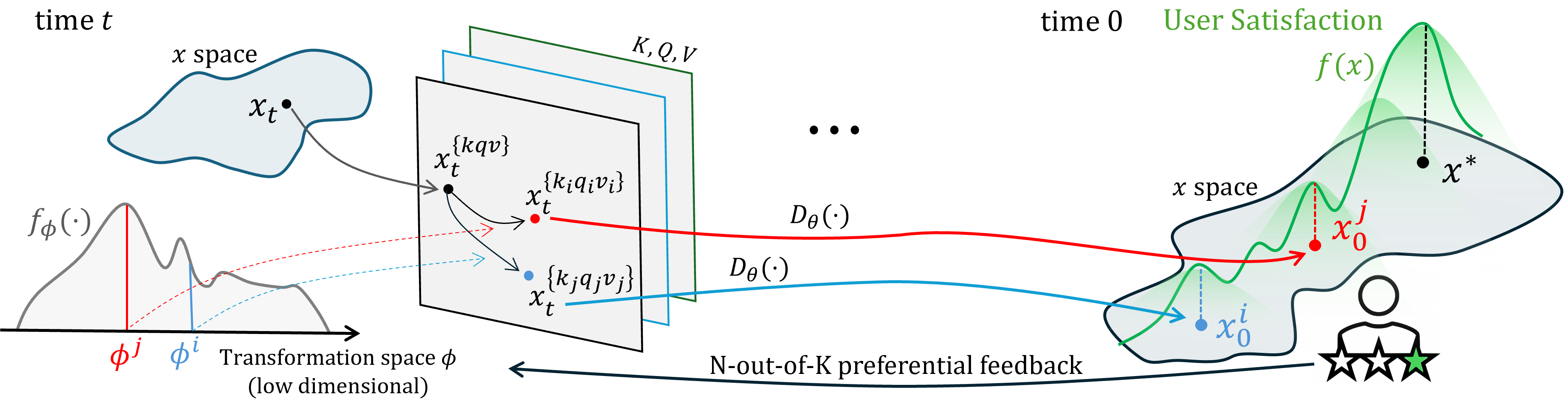}
    \caption{The flow of ideas in {\name}: The BO optimization presents the user with $K$ images and the user chooses $N$ out of $K$ images based on closeness to user's imagined image $x^*$. BO accepts the $N$-out-of-$K$ preferential user feedback and optimizes on the space of transformations applied to the self-attention $KQV$ features, to generate next round of $K$ images, iteratively moving closer to $x^*$. }
    \label{fig:conceptual}
\end{figure*}

We are not the first to utilize human preference in diffusion models; a rich body of work in the areas of \textit{reward-based alignment}, \textit{preference alignment}, and \textit{reinforcement learning using human feedback (RLHF)} is actively exploring this space of ideas \cite{xu2023imagereward, wallace2024dpo, yeh2024training, song2021score}.
Section \ref{sec:related-work} discusses them in detail with the closest being DEMON \cite{yeh2024training}, a training-free inference-time approach that optimizes the noise at each $t$ without back propagation,  using a stochastic optimization by leveraging Probability Flow ODE (PF-ODE) \cite{song2021score}. 
Our main departure from past work lies in upgrading user preference to a multi-choice format, allowing the user to select any subset from $K$ image options.
This offers much richer information, but absorbing this information into blackbox optimization requires a redesign of the likelihood model and the acquisition function.
We conduct this redesign and demonstrate that {\name}'s image generation gets tailored to each individual, is not encumbered by reward-hacking, and does not need offline training using large datasets.
Our users have to indeed wait between each round of feedback, but assuming that is tolerable, the generated image better approaches the user's imagined image, $x^*$.
We sketch our core ideas below.

In standard diffusion, $D_\theta(.)$ denoises a noisy image $x_t$ to $x_0$.
Our goal is to arrive to $x^*$ by performing a human-in-the-loop optimization on $x_t$ as follows:

\begin{equation} 
\underset{v}{\arg \min} ~f(x^*) - f(D_\theta(x_t + v)) 
\end{equation}

Here $f$ is the user's satisfaction function, maximized at $x^*$.
Unfortunately, $f(.)$ is unknown---a blackbox function---hence there is no gradient to be obtained. 
However, it is possible to sample $f(.)$, meaning that we can query the user for feedback for any given image $D_\theta(x)$. 
Using such feedback, we can adopt existing Bayesian optimization (BO) methods using Gaussian Process Regression (GPR), explained in Sec. \ref{sec:background}.
Briefly, BO will propose judiciously chosen variants of $x_t$, say $x_t^j$, and the user will rate how close $D_\theta(x_t^j)$ is to $x^*$.
BO can utilize this feedback to update its predictions, eventually finding 
$\hat{x}_t^*$ such that $D_\theta(\hat{x}_t^*)$ is arbitrarily close to $x^*$. 
Of course, to reduce user burden, the number of user queries needs to be limited to a budget $B$.

Problems arise in realizing this high level idea. 
\circled{1} Operating BO in the high dimensional pixel space is very difficult.
The optimization must be re-cast to a much lower dimensional space, while ensuring that the manifold of correct images, at time $t$, can be reached from $x_t$.
\circled{2} Giving a numerical feedback, $f(D_\theta(x_t^j))$, is known to be difficult for users \cite{tsukida2011analyze} because they may not be able to \textit{quantify} how much worse a given $D_\theta(x_t^j)$ is compared to $x^*$. 
However, it is easier to express preferences between pairs or sets of $K$ images. A user-friendly solution must design the BO framework in a suitable low-dimensional space, incorporating the user's preferential feedback.

Past work have made progress where diffusion models leverage human preference.
For example, \cite{xu2023imagereward, fan2023dpok, wallace2024dpo} train (global) reward models, where offline volunteers express preferences between pairs of images; this reward model becomes a proxy for human preference, which ultimately guides the denoising vector. 
Similar ideas exist in blackbox optimization through preferential likelihood models \cite{chu2005preference}, where GPR accommodates pairwise preferential feedback from users (more in Sec. \ref{sec:background}). 
However, when the function domain is high dimensional, and when a user's preference is pairwise, mapping the function reasonably well incurs excessive user feedback.
Reducing the user burden calls for richer preferential information, and then integrating them into a lower dimensional space for optimization speed up.

\textbf{\textit{Key Contributions:}}
{\name} contributes by observing that \textit{multi-choice preference queries} bring far richer information to the optimization, and shows that such N-out-of-K feedback can be mathematically accommodated by GPR through an updated likelihood model and acquisition function. 
Moreover, building on the empirical success of \cite{nam2024dreammatcher}, {\name} proposes to perform the GPR optimization in a low-dimensional transformation space, where the transformations are applied on $\langle K, Q, V \rangle$ matrices in the attention layer of a diffusion model.
The family of transformations are parameterized by $\phi$ in suitably low dimensions  (see Figure. \ref{fig:conceptual}), but offer flexibility to explore the image manifold around $x_t$. 
In sum, {\name} utilizes multi-choice user feedback to optimize image representations inside the attention layer, empowering users to generate images close to their imagination.

Experiments are designed with a human picking a target image $x^*$, and a starting image $x^{p*}$ that bears similarity to $x^*$ (i.e., language prompts cannot easily convert $x^{p*}$ to $x^*$).
Five baselines attempt to optimize towards $x^*$, either with real human feedback, or with their respective (pre-trained) reward models.
At the end, each baseline submits their final image $\hat{x}^*$ and $30$ external volunteers evaluate the closest match (and other opinion scores).
We also use quantitative metrics and a variety of ablations to shed light on the internal pros and cons of each method, and our own design choices.
Results show that {\name} robustly outperforms other methods and there is room for further improvement. 
\vspace{-0.05in}

%% file: tex/_2_background.tex

\section{Background}
\label{sec:background}
$\blacksquare$ \textbf{Bayesian Optimization}

Bayesian optimization (BO) \cite{frazier2018tutorial,wang2020intuitive} is a non-parametric black-box optimization method that consists of two modules:

(1) {\em Gaussian Process Regression (GPR)}: constructs a probabilistic surrogate of $f$ using the Gaussian likelihood model. The GPR posterior captures our beliefs about the unknown objective function.
Given a set of observations $(\mathcal{X},\mathcal{F})$ and the Gaussian prior kernel, $\mathbf{K}$, the GPR posterior is defined as,
\begin{equation}
\begin{aligned}
P(\mathcal{F}|\mathcal{X}) \sim \mathcal{N}(\mathcal{F}|\boldsymbol\mu,\mathbf{K})
\end{aligned}
\label{eqn:posterior}
\end{equation}

where, $\boldsymbol\mu$ = $\{\mu(x_1),\mu(x_2),\dots,\mu(x_K)\}$ is the best model of the function $f$ given the observations $(\mathcal{X},\mathcal{F})$ and $\mathbf{K}_{ij}$=$k(x_i,x_j)$, $k$ represents a kernel function, the uncertainty map of $f$ over unsampled regions of the input space.

To make predictions $\hat{\mathcal{F}}=f(\hat{\mathcal{X}})$ at new points $\hat{\mathcal{X}}$, GPR uses the current posterior $P(\mathcal{F}|\mathcal{X})$ to define the joint distribution of $\mathcal{F}$ and $\hat{\mathcal{F}}$, $P(\mathcal{F},\hat{\mathcal{F}}|\mathcal{X},\hat{\mathcal{X}})$ as,

\begin{equation}
\begin{aligned}
\begin{bmatrix}
    \mathcal{F} \\ \hat{\mathcal{F}}
\end{bmatrix} \sim \mathcal{N}\left(\begin{bmatrix}
    \mu(\mathcal{X}) \\ \mu(\hat{\mathcal{X}})
\end{bmatrix}, \begin{bmatrix}
    \mathbf{K} & \hat{\mathbf{K}} \\ \hat{\mathbf{K}}^T & \hat{\hat{\mathbf{K}}}
\end{bmatrix}\right)
\end{aligned}
\label{eqn:predict}
\end{equation}

where, $\mathbf{K} = k(\mathcal{X},\mathcal{X})$, $\hat{\mathbf{K}} = k(\mathcal{X},\hat{\mathcal{X}})$, $\hat{\hat{\mathbf{K}}} = k(\hat{\mathcal{X}},\hat{\mathcal{X}})$ and $(\mu(\mathcal{X}), \mu(\hat{\mathcal{X}})) = \mathbf{0}$.

The conditional distribution and hence prediction of $\hat{\mathcal{F}}$ from the joint distribution is, 

\begin{equation}
\begin{aligned}
P(\hat{\mathcal{F}}|\mathcal{F},\mathcal{X},\hat{\mathcal{X}}) \sim \mathcal{N}(\hat{\mathbf{K}}^T\mathbf{K}^{-1}\mathcal{F},\hat{\hat{\mathbf{K}}} - \hat{\mathbf{K}}^T\mathbf{K}^{-1}\hat{\mathbf{K}})
\end{aligned}
\label{eqn:predict-condt}
\end{equation}

The proof and explanations of all the above are clearly presented in \cite{wang2020intuitive}).

(2) {\em Acquisition function (ACF)}: a sampling strategy that seeks to identify future observations that would improve the likelihood model. The goal, is to use evidence (observations) and prior knowledge (Kernel) to maximize the posterior at each step, so that each new evaluation decreases the distance between the true global maximum and the expected maximum given the GPR model.
We use ``Expected Improvement'' (EI) ACF in this work. Given previous observations $(\mathcal{X},\mathcal{F})$, let $f^* = \max_{x \in \mathcal{X}}f(x)$ be the current function minimum (i.e., the maximum observed till now). 
If a new observation $f(x')$ is made at $x'$, then the new maximum will be either $f(x')$ if $f(x') \geq f^*$ or $f^*$ if $f(x') \leq f^*$.

Hence, the improvement from observing $f$ at $x'$ is \\
\[[f(x') - f^*]^+.\] where, $a^+ = \max(a,0)$.

We want to choose $x'$ that maximizes this improvement. 
However, $f(x')$ is unknown until the observation is made, so we choose $x$ that maximizes the expectation of this improvement. 
\emph{Expected Improvement} is thus defined as:
\begin{equation}
\begin{aligned}
\textbf{EI}(x|\mathcal{X},\mathcal{F}) = \mathbf{E}[[f(x) - f^*]^+|\mathcal{X},\mathcal{F}]
\end{aligned}
\end{equation}
where, $\mathbf{E}[\cdot|\mathcal{X},\mathcal{F}]$ is the expectation taken on the GPR posterior distribution (Eqn. \ref{eqn:posterior}) given observations $(\mathcal{X},\mathcal{F})$. Thus, the next sample to make an observation at is:
\begin{equation}
\begin{aligned}
x_{\textbf{EI}} &= \argmax_{x \in \mathbf{R}^D}\textbf{EI}(x_i|\mathcal{X},\mathcal{F}) \\ &= \argmax_{x \in \mathbf{R}^D} \mathbf{E}[[f(x)-f^*]^+|\mathcal{X},\mathcal{F}]
\end{aligned}
\label{eqn:ei}
\end{equation}

Conventional Bayesian Optimization requires that each function evaluation have a scalar response. However, in applications requiring human judgement, preferences are often more accurate than ratings. The scenario we consider in this work is about presenting two or more samples to a user and requiring only that they indicate preference. This means we cannot query the objective function directly. In this case, $f$ is considered a latent unobservable function, to be inferred from the preferences, which allows for the use of Bayesian optimization approach \cite{brochu2010tutorial}.

$\blacksquare$ \textbf{Preferential Bayesian Optimization}

Let us first consider the case of paired preference data i.e., the user is presented two choices to pick one from. A user's preference information (favoring A over B) can be used to construct a likelihood model.
The \emph{probit} likelihood model \cite{thurstonelaw,mosteller1951remarks} allows us to infer $f$ from the binary preference observations. 
Assume we have shown the user $M$ pairs of data from a set of $N$ samples $ [x_1,\dots,x_N]$. The data set
therefore consists of the ranked pairs:
\begin{equation}
\begin{aligned}
\mathcal{X} = \{a_i \succ b_i; i=1,\dots,M\} \quad a_i,b_i \in \mathcal{Y}
\end{aligned}
\end{equation}
Assuming noisy observations, the latent function values are, 
\begin{equation}
\begin{aligned}
v(a_i) = f(a_i) + \delta, \quad v(b_i) = f(b_i) + \delta,
\end{aligned}
\label{eqn:blt}
\end{equation}
where, $\delta \sim \mathcal{N}(0,\sigma^2)$ is Gaussian noise.
Following the GPR modeling of $f$ (Eqn. \ref{eqn:posterior}),  The Bradley-Teller-Luce (BTL) \cite{stein1999interpolation} model defines the probability that data $a$ is preferred to data $b$ as,
\begin{equation}
\begin{aligned}
P(a_i &> b_i | f(a_i),f(b_i)) \\
&= \int \int P(a_i > b_i | f(a_i) + \delta_a,f(b_i) + \delta_b) \\
& \quad \cdot \mathcal{N}(\delta_a;0,\sigma^2)\mathcal{N}(\delta_b;0,\sigma^2) d\delta_a d\delta_b \\
&= \Phi(z_i)
\end{aligned}
\label{eqn:pbo-likelihood}    
\end{equation}
where $z_i = \frac{f(a_i)-f(b_i)}{\sqrt{2}\sigma}$ and $\Phi(z) = \int_{-\infty}^{z} \mathcal{N}(\gamma;0,1) d\gamma$ is the Gaussian CDF. This preferential likelihood model is called a \textbf{Binomial-Probit} regression model. It can be converted to a \textbf{Binomial-Logit} model by setting $\varphi(z_i) = \frac{1}{1+\exp(-z_i)}$. 

Given $M$ pairwise observations, The BTL logit likelihood model of observing the preference relations given the latent function values $f(x_i)$ is,
\begin{equation}
\begin{aligned}
P(\mathcal{X}|\mathbf{f}) = \prod_{k=1}^M P(a_i > b_i | f(a_i),f(b_i))
\end{aligned} 
\end{equation}
and the GPR posterior is,

\begin{equation}
\begin{aligned}
P(\mathbf{f}|\mathcal{X}) = \frac{P(\mathbf{f})}{P(\mathcal{X})} \prod_{k=1}^M P(a_i > b_i | f(a_i),f(b_i))
\end{aligned}
\label{eqn:pbo-posterior}
\end{equation}
where $P(\mathbf{f})$ is the prior, and $P(\mathcal{X}) = \int P(\mathcal{X}|\mathbf{f})P(\mathbf{f})d\mathbf{f}$.
The detailed proofs are well established in \cite{chu2005preference} and more details are in Appendix \ref{sec:appendix_pairwise}. 

 This likelihood serves as the pathway for probabilistic modeling of human preference and its optimization using the Bayesian framework. Thus, BO applied to user preference data 
 is Preferential BO (PBO) \citep{chu2005preference}.

$\blacksquare$ \textbf{Latent Diffusion Models (LDMs)}

Diffusion models generate images via a reverse denoising process. The forward process adds noise to a clean data point $x_0$ using scales $\bar{\alpha}_t = \prod_i^t \alpha_i$, yielding $x_t = \sqrt{\bar{\alpha}_t} x_0 + (\sqrt{1 - \bar{\alpha}_t})\epsilon$, where $\epsilon \sim \mathcal{N}(0, I)$ and $x_T \sim \mathcal{N}(0,I)$.

The reverse process recovers the image by modeling $p_{\theta}(x_{t-1}|x_t)$. This relies on estimating the clean sample $\hat{x}_0$ using Tweedie’s formula and a trained denoiser $\epsilon_{\theta}$:
\begin{equation}
    \hat{x}_0 = \frac{x_t - \sqrt{1-\bar{\alpha}_t}\,\epsilon_\theta(x_t, t)}{\sqrt{\bar{\alpha}_t}}
\end{equation}

LDMs improve efficiency by running this process in a VAE-compressed latent space. They enable control (e.g., text-to-image) via conditional denoisers $\epsilon_\theta(x_t, c, t)$, where a domain encoder $\tau_{\theta}$ projects prompts $c$ (e.g., CLIP embeddings) for the denoiser.

$\blacksquare$  \textbf{Attention in Diffusion Models}

The denoiser network of Diffusion models utilize attention mechanism to prioritize relevant features for consistency and control. Cross-attention aligns visual features with conditioning input $c$ (text), while self-attention models spatial dependencies and global correlations to ensure semantic integrity.

Both mechanisms project image features at time $t$ into query ($Q_t$), key ($K_t$), and value ($V_t$) vectors, yielding the attention map:
\begin{equation}
    \text{Attention}(Q_t,K_t,V_t) = \text{Softmax}\left(\frac{Q_tK_t^T}{\sqrt{d}}\right)V_t
\end{equation}
where $Q_t,K_t,V_t \in \mathbb{R}^{H\times W \times d}$ denote projections across height $H$, width $W$, and channels $d$.

$Q$ and $K$ dictate spatial attributes \cite{nam2024dreammatcher}. In cross-attention, $V$ connects these to conditioning tokens, whereas in self-attention, $V$ controls non-spatial features (color, texture), enabling fine-grained generation control.

%% file: tex/_3_method.tex
\begin{figure}[htbp]
\centering
\includegraphics[width=1\columnwidth]{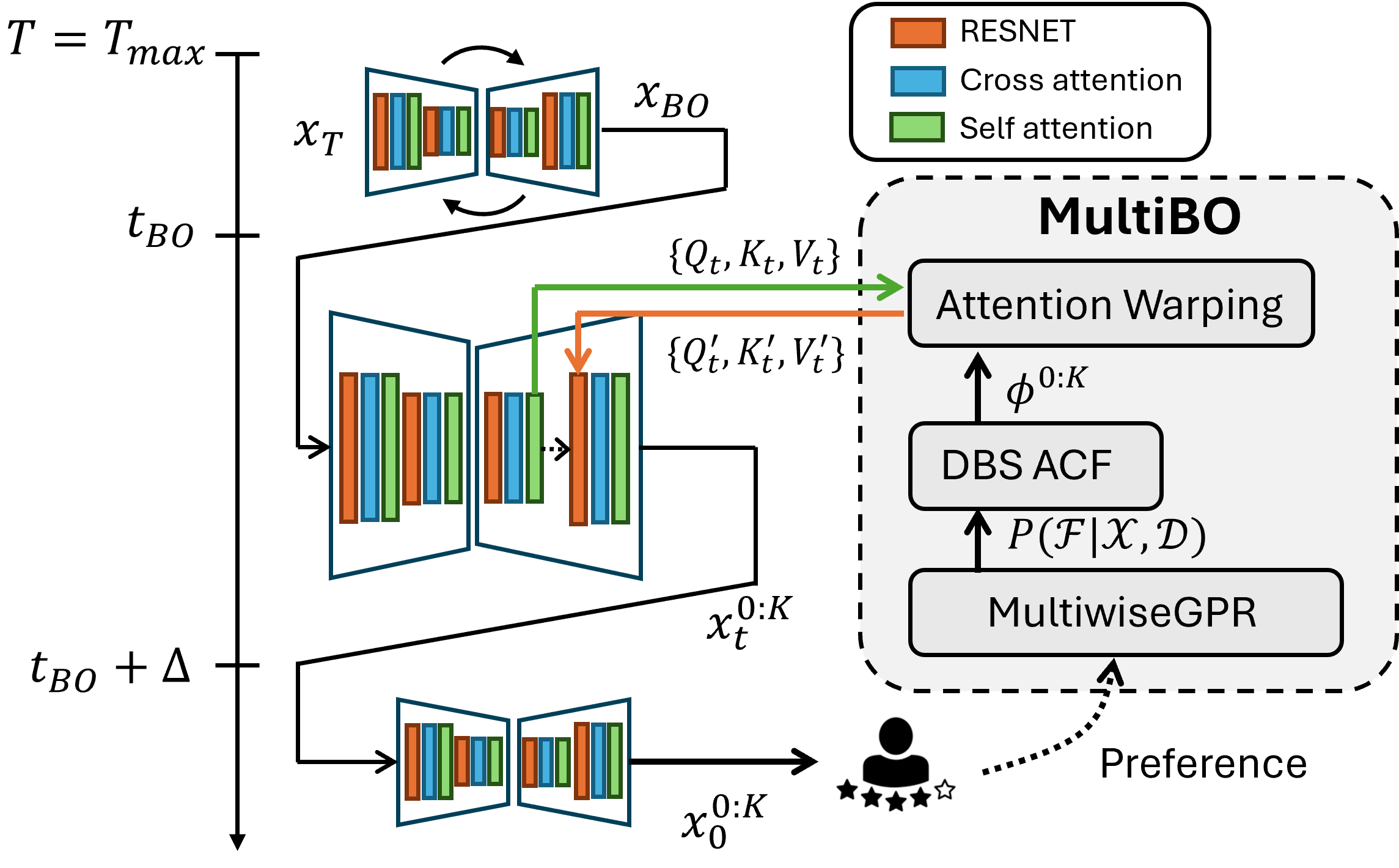}
\caption{\small {\name} Image Personalization Pipeline. {\name} optimizes the self attention $Q,K,V$ features at time interval $[t_{\text{BO}}, t_{\text{BO}} + \Delta]$ over warping transformation space $\mathcal{Y}$. At each iteration $i$, {\name} offers $K_i$ transform parameter choices and the user picks the $N$ ``best'' option(s) from the corresponding $K$ attention-modified images. \textbf{The MultiwiseGPR} likelihood models the unobservable user \emph{satisfaction} function $f$ from the user preferences and the \textbf{Dynamic Balanced Subspace (DBS)} acquisition function prescribes the next set of $K_{i+1}$ warp parameters.}
\label{fig:outline}
\end{figure}

\section{{\name}}
Briefly, we aim to adjust diffusion generation $x_0 = D_\theta(x_t)$ to closely match a target $x^*$.
We cast this as a human-in-the-loop Black-box optimization (BBO) problem and use Bayesian Optimization (BO) approaches. As discussed in Sec. \ref{sec:background}, Preferential Bayesian Optimization (PBO) provides the framework for taking in user feedback in the form of paired preference and modeling unknown \emph{user satisfaction} function $f$ using Pairwise GPR (Eqn. \ref{eqn:pbo-likelihood}). Optimizing the corresponding GPR posterior (Eqn. \ref{eqn:pbo-posterior}) produces $\hat{x}^*$.   

Two challenges arise in our problem: (1) we can only obtain finite user feedback (typically $50$ questions). Therefore, each user preference feedback has to reveal as much information as possible about unknown $f$. (2) BO has to operate on very high-dimensional RGB Pixel Space to find $\hat{x}^*$, unsuitable for BBO.   

In {\name} we propose two modules: \textbf{\textit{Multi-choice Preferential Bayesian Optimization}} and \textbf{\textit{Self-Attention Warping}} to address these challenges, as illustrated in Figure \ref{fig:outline}.


\begin{algorithm}[H]
\caption{Personalized Image Generation with {\name}}
 \label{alg:multibo}
\KwIn{
Prompt $P$ , denoiser $\epsilon_{\theta}$, $x_T$ image sample seed, optimization budget $B$, Edit time $t_{\text{BO}}$ and interval $\Delta$, BO hyperparameters $\gamma$, $K$ image choices $\mathcal{Z}$, User preference $g: x_w > x_{\mathcal{Z}/w}, w \in \{1,\dots,K\}$.
}
\KwOut{Best image sample $\hat{x}^*$ after $B$ iterations}

\BlankLine
\For{$i = 0,\dots,B$}{
   \If{i = 0}{
   Initialize $N_0$ sets of $K$ random warping parameters $\mathcal{X}_0 = (\Theta^{(j)}_1,\Theta^{(j)}_1,\dots,\Theta^{(j)}_{K}), j=0,\dots,N_0$\\
   }
   \Else{   
           $\Theta^{(i)}_{1:K} = \text{DBS}(g, P(\mathbf{f}|\mathcal{X}_i))$ \tcp*[r]{Next set of choices}
           $\mathcal{X}_i = [\mathcal{X}_{i-1},\Theta^{(i)}_{1:K}]$ \tcp*[r]{Aggregated data}
       }
\greenbox{\textbf{Self Attention Editing:\\}\For{$t = t_{\text{BO}}, t_{\text{BO}} + \Delta$}{
   $Qs_t, Ks_t, Vs_t \leftarrow \epsilon_{\theta}(x_t,t,P)$
   
    $\{Qs'_t, Ks'_t, Vs'_t\}_{1:K} = \text{Warp}(Qs_t, Ks_t, Vs_t;\Theta^{(i)}_{1:K})$

   $\epsilon_{1:K} =  \epsilon_{\theta}(x_t,t,P; \{Qs'_t, Ks'_t, Vs'_t\}_{1:K})$
   
   $\{x_0\}_{1:K} = \text{Sampler}(x_t,t,P, \epsilon_{1:K})$}}
   
   $g: x_{0,w} > x_{0,\mathcal{Z}/w}$ \tcp*[r]{Obtain User preference choice}
   
   Fit GPR Posterior $P(\mathbf{f}|\mathcal{X}_i)$ (Eqn. \ref{eqn:multi-posterior})}
   
   $Qs^*_t, Ks^*_t, Vs^*_t = \text{Warp}(Qs_t, Ks_t, Vs_t;\Theta^*)$ \tcp*[r]{Best params},   
   
   $\epsilon^* =  \epsilon_{\theta}(x_t,t,P; Qs^*_t, Ks^*_t, Vs^*_t)$,
   
   Best Image: $\hat{x}^* = \text{Sampler}(x_t,t,P, \epsilon^*)$

\end{algorithm}

\subsection{Multi-choice Preferential BO (Multi-choicePBO)}
In PBO, for the PairwiseGPR model to be a good approximation for $f$, many observations (preference feedback) are required. An observed pair only provides information about $f$ (resolves uncertainty) for two halves of the input space $\mathcal{Y}$ and more data is needed to sufficiently map the function landscape. This is further exacerbated by increasing dimensionality of the input space. Large number of queries will put considerable burden on the user, diminishing feasibility for human-in-the-loop setting.

To reduce this user burden is to increase the information extracted from each user preference.
One way to achieve this is to expand the choice set to $K$ images. This makes intuitive sense, as each observation now resolves $f$ on $K$ partitions of the input space $\mathcal{Y}$. The user has the flexibility to choose $N$-out-of-$K$ best images, $N > 1$ if the user thinks they are equally good or they contain complementary aspects related to the target image. This can significantly bring down the number of user queries at the expense of slight increase in user's cognitive load of selecting $N$-out-of-$K$ images instead of $1$-out-of-$2$. 

To handle this complex preference signal, we design \textbf{\textit{MultiwiseGPR}}, a likelihood model that maps multi-choice preference relations to $f$ and \textbf{\textit{Dynamic Balanced Subspace (DBS)}} acquisition function that supplies the next set of $K$ choices based on the MultiwiseGPR posterior.

$\blacksquare$ \textbf{MultiwiseGPR}

We take the discussion on the more general $N$-out-of-$K$ setting to the Appendix \ref{sec:appendix_subset}. Consider the special case when $N=1$ i.e., user picks 1 image out of $K$ choices. Consider a set of $N$ distinct samples $x_i \in \mathcal{Y} \subseteq \mathbb{R}^D$; 
$[x_i : i = 1,..., N]$. Let $\mathcal{Z}$ denote the $K$ choices chosen from it for one observation set. \\
The set of $M$ observed multiwise preference relations on the choice set $\mathcal{Z}_i, i=1,\dots,M$ is,
\begin{equation}
\begin{aligned}
\mathcal{X}_i = \{a_i^{(w)} \succ \{a_i^{(j)}\}_{j \in  \mathcal{Z}_i/w} : i = 1,..., M\}
\end{aligned}
\end{equation}
where $\{a_i\}_{\mathcal{Z}_i} \in [x_1,\dots,x_N]$ 
and $a_i^{(w)} \succ \{a_i^{(j)}\}_{j \in  \mathcal{Z}_i/w}$ means $a_i^{(w)}$ is the preferred winning sample over $K-1$ other choices in choice set $\mathcal{Z}_i$. This preference relationship is modeled by a polycotomous regression model \cite{held2006bayesian} -- multinomial-logit regression model.

\textit{\textbf{Multinomial-Logit Regression}}: 
Let us consider the simple case when $M=1$ observation. There are $K$ samples (images) in the choice set $\{a_1,\dots,a_K\} \in \mathcal{Z}$. We use a Gaussian prior, and the corresponding latent function values are $\mathbf{f} = (f(a_1),\dots,f(a_K))$. 

If we assume IID Gaussian noise $\mathcal{N}(0,\sigma)$, then, following Eqn. \ref{eqn:blt},
\[v_j = f_j + \delta_j, \delta \in \mathcal{N}(0,\sigma) \]
Picking choice $a_i^{(w)}$ out of $\{a_i^{(j)}\}_{j \in \mathcal{Z}_i},   i = 1,\dots,M$ is modeled  by a categorical distribution -- multinomial-probit (multi-choice version of the binomial in \eqref{eqn:pbo-likelihood}) as,

\begin{equation}
\begin{aligned}
P(Y = w | \mathbf{f}) = P(v_w = max_j v_j)\\ = \int \mathbf{1}\{f_w +\delta_w \ge f_j +\delta_j,\forall j \neq w\}\Phi(\delta)d\delta
\end{aligned}
\end{equation}

where, $Y$ is picking winning index $w$ in $K$ choices $\mathcal{Z}$. This is a multivariate normal orthant probability (multidimensional CDF). There is no closed-form expression for this multinomial-probit regression model.
Following PBO \cite{chu2005preference}, we replace the probit model for the logistic model. \\
Let's update our assumption to IID $\delta \in \text{Gumbel}(0,1)$ noise in the observation, the multinomial-logit distribution is,

\begin{equation}
\begin{aligned}
P( Y = w | \mathbf{f}) = \frac{\exp(f_w)}{\sum_{j=1}^K\exp(f_j)}
\end{aligned}
\label{eqn:multi-likelihood}
\end{equation}
The logistic distribution is characterized by the \textit{softmax} function (parallels sigmoidal definition in Eqn. \ref{eqn:pbo-likelihood}).
The joint likelihood of observing  $M$ multi-choice observations given the latent function values $f(x_i)$ is the product of the likelihood function of each observation in Eqn. \ref{eqn:multi-likelihood},

\begin{equation}
\begin{aligned}
P(\mathcal{X}|\mathbf{f}) = \prod_{k=1}^M P(Y = w |\mathbf{f})
\end{aligned}
\end{equation}
and the corresponding Multiwise GPR posterior is,

\begin{equation}
\begin{aligned}
P(\mathbf{f}|\mathcal{X}) = \frac{P(\mathbf{f})}{P(\mathcal{X})} \prod_{k=1}^M P(Y = w |\mathbf{f})
\end{aligned}  
\label{eqn:multi-posterior}
\end{equation}
where $P(\mathbf{f})$ is the Gaussian prior, and $P(\mathcal{X}) = \int P(\mathcal{X}|\mathbf{f})P(\mathbf{f})d\mathbf{f}$. The proofs for the likelihood and estimation of the posterior are found in Appendix \ref{sec:appendix_multiwise} and \cite{held2006bayesian}.

Thus, we have a probabilistic mapping between multi-choice preference and $f$ that is leveraged by the acquisition function in finding $\hat{x}^* = \argmax f$.

$\blacksquare$ \textbf{Dynamic Balanced Subspace (DBS)}: 

MultiwiseGPR expects future observations $\mathcal{X}$ that improve the belief of $f$ (Eqn. \ref{eqn:multi-posterior}). The naive way is to extend the Expected Improvement (EI) ACF (Eqn. \ref{eqn:ei}) to $K$-EI i.e., $K$-jointly maximize the posterior expectation. This joint optimization of an already intractable expectation incurs a tremendous computational load (especially when $K$ is large). In human-in-the-loop settings, we design a light ACF offering $K$ ``good'' choices that balance observed preferences (\emph{exploitation}) with diverse alternatives (\emph{exploration}).

DBS ACF bypasses the $K$-EI computational trap by computing only 1 sample, $x_{\text{EI}}$. Armed with $x_{\text{EI}}$ and the the current best sample $\hat{x}^{*}$, DBS ACF constructs a set of $K$ anchor points that act as \textit{bridge vectors} $\mathbf{v}_{bridge}$ connecting them,
\begin{equation}
\mathbf{v}_{bridge, i} = \hat{x}^* + \gamma_i (x_{\text{EI}} - \hat{x}^*), \gamma _{i}\in [0,1]
\end{equation}

The key intuition is that $\mathbf{v}_{bridge}$ modulate the explore-exploit trade-off between BO's forward thrust $x_{\text{EI}}$ and the belief $\hat{x}^{*}$ ensuring that the preference set $\mathcal{Z}$ always includes candidates that represent a direct transition from the best known $\hat{x}^*$ to the theoretical optimal global point $x_{\text{EI}}$.
The final $K$ choices ($x^+)$ presented to the user are constructed by random perturbations $\delta$ along the $\mathbf{v}_{bridge}$ directions,
\begin{equation}
\label{eqn:bridge}
\begin{aligned}
x^+_i = \text{proj}_{\mathcal{Y}} \left( \mathbf{v}_{bridge, i} + \delta_i \right), \quad i=0, \dots, K-1
\end{aligned}
\end{equation}

where $\text{proj}_{\mathcal{Y}}$ is the projection onto the input space $\mathcal{Y} \subseteq \mathbb{R}^D$.

As we are cognizant of the user burden constraints on the optimization especially when the input space $\mathcal{Y} \subseteq \mathbb{R}^D$ is high-dimensional, DBS ACF samples the perturbations $\delta \in \Omega$ from a subspace $\Omega \subseteq \mathcal{Y}$, instead of $\mathcal{Y}$. 
Borrowing from the work Bounce \cite{papenmeier2023bounce} that performs high-dimensional Bayesian Optimization using embedding spaces of increasing dimensionality, we design our DBS ACF to operate over a subspace constructed dynamically as the optimization proceeds.

\begin{algorithm}[H]
\caption{Dynamic Balanced Subspace (DBS) ACF}
\label{alg:dbs}
\KwIn{BO iteration $i$; observed data $\mathcal{X}_i$; Current GPR posterior $P(\mathbf{f}|\mathcal{X}_i)$; Current best sample $\hat{\Theta}^*$; Warp parameter space $\mathcal{Y} \subseteq \mathbb{R}^D$}
\KwOut{$\Theta^+_{0:K-1}$ - next set of $K$ warping parameter choices}
Expected Improvement sample: $x^+ = \text{EI}(P(\mathbf{f}|\mathcal{X}_i|\mathcal{Y}))$\\
Bridge points:\\ $\Theta_{bridge, i} = \hat{\Theta}^* + \gamma_i (\Theta^+ - \hat{\Theta}^*), \gamma _{i}\in [0,1]$
\redbox{\textbf{Dynamic Subspace Selection:}\\
- Compute gradient matrix $\mathbb{C}$ from MultiwiseGP posterior mean $\mu (\mathbf{x})$ near $\hat{\Theta}^{*}$: $\hat{\Theta} = \hat{\Theta}^{*} + \eta $
\begin{center}
    $\mathbb{C} = \frac{1}{N} \sum_{i=1}^N \nabla \mu(\hat{\Theta}) \nabla \mu(\hat{\Theta})^T$
\end{center}
Eigen decomposition: $\lambda$,$\mathbf{u} = \text{EVD}[\mathbb{C}]$\\
Target dimensions: $d = \argmax_j \frac{\lambda_j}{\lambda_{j+1}} \quad j \in \{1,2,\dots, D\}$ (Spectral Gap)\;}
$K$ choices: $\delta_i = \sigma \sum_{j=1}^d \alpha_{i,j} \sqrt{\lambda_j} \mathbf{u}_j \quad \alpha _{i,j} \in [-1, 1]$\\
$\Theta^+_i = \text{proj}_{\mathcal{Y}} \left( \Theta_{bridge, i} + \delta_i \right), \quad i=0, \dots, K-1$
\end{algorithm}

\textbf{\textit{d-dimensional Subspace}}:
We identify the most influential directions in the $D$-dimensional input space $\mathcal{Y}$ by constructing an uncentered covariance matrix $\mathbb{C}$ of the GPR posterior gradients. Let $\mu (\mathbf{x})$ denote the posterior mean of the Multiwise GPR model. We estimate $\mathbb{C}$ by averaging the outer products of the gradients at $H$ design points sampled in the neighborhood of the current best point $\hat{x}^*$ as,
\begin{equation}
\label{eqn:spectral-matrix}
\begin{aligned}
C = \frac{1}{H} \sum_{i=1}^H \nabla \mu(\mathbf{x}_i) \nabla \mu(\mathbf{x}_i)^T, x_i = \hat{x}^* + \eta
\end{aligned}
\end{equation}
We determine the subspace dimension $d$ by the \textbf{Spectral Gap}, following the eigendecomposition of $\mathbb{C}$,
\begin{equation}
\begin{aligned}
d = \argmax_i \frac{\lambda_i}{\lambda_{i+1}} \quad i \in \{1,2,\dots, D\}
\end{aligned}
\label{eqn:spectral_gap}
\end{equation}
This identifies the threshold $d$ after which additional dimensions contribute significantly less to the visual change of the image, allowing DBS ACF to adaptively \textit{unlock} more dimensions when the function $f$ landscape is complex and \textit{collapse} to a lower-dimensional space when a few dimensions dominate the visual features.
Hence, the $d$-dimensional subspace $\Omega$ represents a $d-$volume in the $D$-dimensional input space that contains the most visually significant dimensions.
The perturbations $\delta$ in Eqn. \ref{eqn:bridge} are now samples from the $d-$Subspace $\Omega$, 
\begin{equation}
\begin{aligned}
 \delta_i = \sigma \sum_{j=1}^d \alpha_{i,j} \sqrt{\lambda_j} \mathbf{u}_j   
\end{aligned}
\label{eqn:perturb}
\end{equation}
where, each coordinate $\alpha _{i,j} \in [-1, 1]$ is  \textit{Eigen-Weighted} by scaling with corresponding eigenvalue $\lambda _{j}$. This ensures that perturbations along the most sensitive dimensions are more pronounced than those along dimensions with lower influence.

The eigen weighting and perturbations along the bridge vector ensure that the $K$ choice set of $\mathcal{Z}$ does not collapse into a redundant line between the anchors $\{\hat{x}^*, x_{\text{EI}}\}$, i.e., even if $\mathbf{x}_{\text{EI}}$ and $\hat{x}^{*}$ are close in the input space, the user is presented with diverse visual variations spanning the most influential latent dimensions of the diffusion model space.

Multi-choice Preferential Optimization provides the framework for translating multi-choice preference data to latent ``user satisfaction" function $f$ and optimizing $f$ to ultimately produce $\hat{x}^* = \argmax f$.

\subsection{Self-Attention Warping}
Theoretically, Multi-choicePBO would operate on diffusion's latent space $x_t \in \mathbb{R}^D$. However, this is practically infeasible as the latent space dimensions $D$ are $\approx 16k$. 
Attention mechanism in diffusion influences both local and global semantic and structural attributes of a generated image. Thus, Multi-choicePBO optimization on the attention features offers the unique opportunity to affect both global and local changes without the dimensionality cost. 
As the user has already constructed the most expressive prompt, cross-attention features are not relevant to the task at hand. Instead, we optimize on self-attention $Q,K,V$ features that controls spatial features of the image as well as attributes like texture, shape, color, etc. 

Keeping the practical constraints of human-in-the-loop optimization in mind, we are interested in further constraining the Multi-choicePBO optimization in $Q,K,V$ space. A reasonable constraint design restricts optimization to valid transformations of the attention space. Since the attention mechanism directly correlates with the pixel-level features of the image, we employ a family of transformations typically applied to images -- Warping \cite{truong2021warp}. 

A \textit{\textbf{Warping}} transform is a functional mapping of each pixel from a reference image to a transformed image. This parallels {\name}'s optimization goal of finding that transformation that maps $x_0$ to the target $x^*$. 

Affine transformations are linear and provide global alignment; while Thin Plate Spline transforms (TPS) are non-linear deformations that enable local refinement.

\textit{Affine}: $\mathbf{x}' = A\mathbf{x} + \mathbf{\tau}$
where $A\in\mathbb{R}^{2\times2}\) and \(\mathbf{\tau}\in\mathbb{R}^2$.

\textit{Thin-Plate Spline (TPS) Transformation}:
$f(\mathbf{x}) = a_1 + a_x x + a_y y + 
\sum_{i=1}^N w_i U(\|\mathbf{x}-\mathbf{c}_i\|),$
with $U(r) = r^2 \ln(r^2)$.

{\name} warps self-attention $Q,K,V$ features using a composed transform of affine and TPS, 

\textit{Affine + TPS Composition}: $I_q^{\text{warp1}} = \text{Affine}(I_q),\qquad
I_q^{\text{warp2}} = \text{TPS}(I_q^{\text{warp1}}).$

Thus, {\name}'s attention optimization problem is,
\begin{equation}
\begin{aligned}
Qs^*_t, Ks^*_t, Vs^*_t &= \argmax_{\Theta \in \mathcal{Y}} &\text{W}(Qs_t, Ks_t, Vs_t;\Theta) \\ & \text{BO-iters} \leq B
\end{aligned}
\label{eqn:opti-attn}
\end{equation}
where, $\textbf{W}(\cdot)$ is the warping function, the constrained optimization search space, $\mathcal{Y}$ is the Affine+TPS warping space rather than $ \mathbb{R}^{H\times W \times d}$ and $t$ is the diffusion timestep. 
The optimization framework is Multi-choicePBO. The image personalization pipeline of {\name} is presented in Algorithms \ref{alg:multibo}, and \ref{alg:dbs}.

%% file: tex/_4_results.tex
\section{Experiments}

\begin{figure*}[t]
    \centering
    \includegraphics[width=1.9\columnwidth]{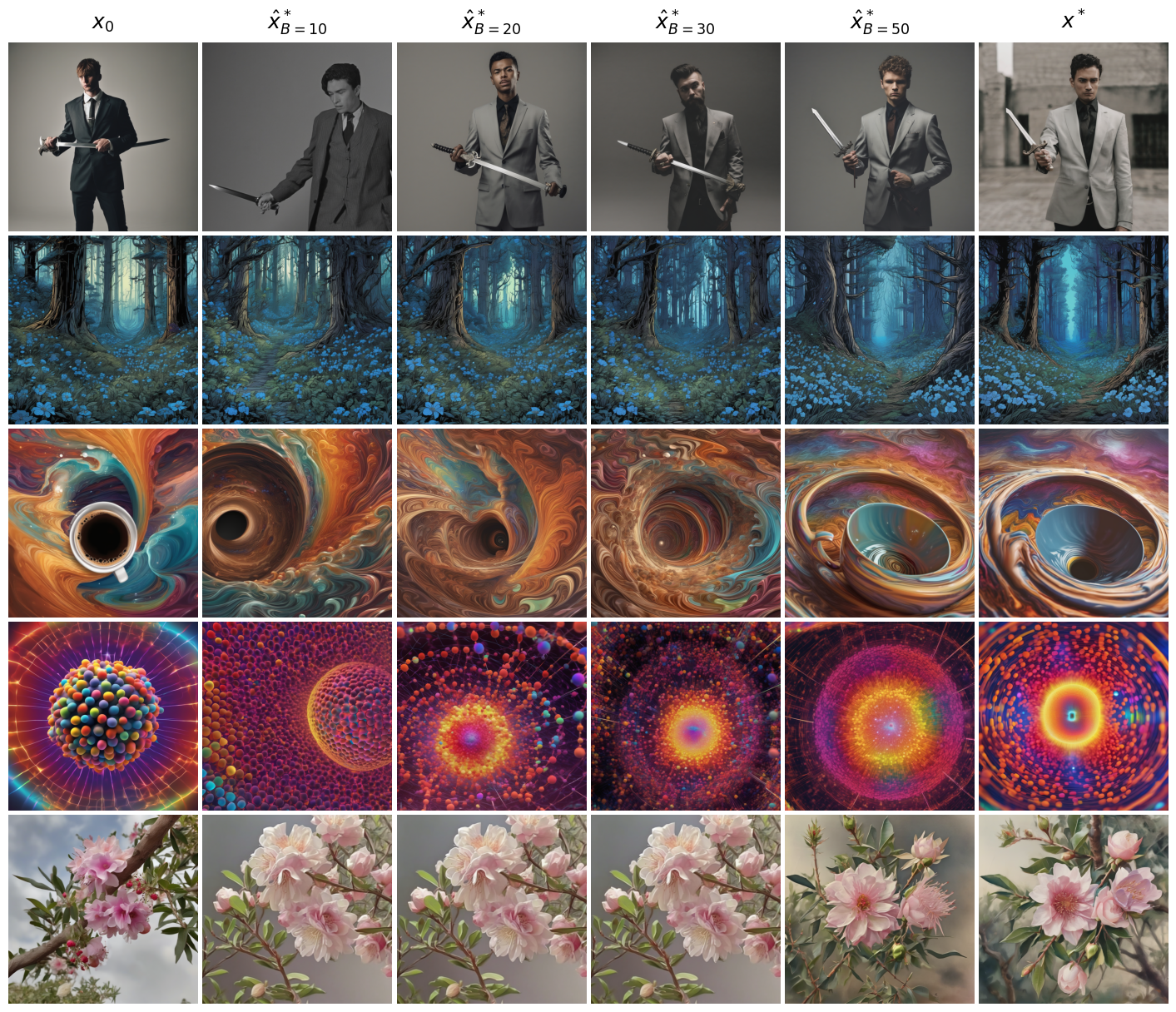}
    \caption{Qualitative Results-- {\name} optimization progress: Starting image $x_0 = D_{\theta}(x_t)$ after prompting, best image $\hat{x}^*$ after $B=10,20,30,50$ iters, and the true target $x^*$. For prompts : \textit{``A person in a suit holding a sword.", ``A forest with blue flowers illustrated in a digital matte style by Dan Mumford and M.W Kaluta.", ``A swirling, multicolored portal emerges from the depths of an ocean of coffee, with waves of the rich liquid gently rippling outward. The portal engulfs a coffee cup, which serves as a gateway to a fantastical dimension. The surrounding digital art landscape reflects the colors of the portal, creating an alluring scene of endless possibilities.",``an electron cloud model is displayed in vibrant colors with a light spectrum background, showcasing the probability distribution of electrons around the nucleus. the image resembles digital art with pixelated elements, bringing a modern, educational twist to atomic structure visualization.", ``The fragrant flowers bloomed on the sturdy stem and the thorny bush."}.}
    \label{fig:multi_iters}
    \vspace{-1.7em}
\end{figure*}

$\blacksquare$ \textbf{Datasets}:

We curate prompts and corresponding target images $x^*$ from popular prompt datasets in diffusion image editing space like AttendExcite \cite{ae}, T2ICompBench \cite{huang2023t2i}, RareBench \cite{park2025raretofrequent} and preference alignment space such as HPSv2 \cite{wu2023humanpreferencescorev2} benchmark dataset, GenEval \cite{ghosh2023geneval}, PartiPrompts \cite{yu2022scaling}, Pick-a-Pic \cite{kirstain2023pick}, and Dalle \cite{zhang2024itercomp}. 

$\blacksquare$ \textbf{Baselines}: 

(1) \emph{Preference Alignment}: We compare with training-based methods: DiffusionDPO \cite{wallace2024dpo} and IterComp \cite{zhang2024itercomp}, training-free methods: DNO \cite{tang2024inference}, DAS \cite{kim2025test}, and DEMON \cite{yeh2024training}. \\
(2) \emph{{\name}$_{<\text{reward}>}$}: Replace the human scorer with popular reward scores and target-based metrics, CLIP-I2I and LPIPS. This category of baselines is closest to the alignment works except with a different optimization (BO) algorithm. \\
(3) \emph{L2-guided}: the ideal upper limit, constructed by directly guiding diffusion denoising with classifier guidance in the form of L2 loss w.r.t target $x^*$.  \\
(4) \emph{{\name}$_{\text{Subspace}}$}: We replace attention space optimization with {\name} applied on a $100D$ subspace of $x_t$ (Following \cite{chen2024exploringlowdimensionalsubspacesdiffusion}).

$\blacksquare$ \textbf{Metrics}:

We compare {\name} against popular preference alignment reward metrics like PicScore \cite{kirstain2023pick}, HPSv2 \cite{wu2023humanpreferencescorev2}, Aesthetic \cite{schuhmann2024laion}, ImageReward \cite{xu2023imagereward}, and VILA \cite{ke2023vila}.
For target alignment, we use CLIP I2I, and LPIPS w.r.t $x^*$.

$\blacksquare$ \textbf{Implementation Details}:

\textbf{\textit{Diffusion Parameters}}: We implement {\name} on SDXL \cite{sdxl} with DDIM for $50$ inference steps and guidance scale $=5.0$. {\name} performs attention editing during the first $20\%$ of the denoising steps. 

\textbf{\textit{Attention Parameters}}: We target the middle to later attention layers (the decoder layers in the case of UNets) as they have been shown empirically in \cite{liu2024towards} to produce significant spatial changes without the loss of consistency. Nine grid points define the TPS transformation \cite{truong2021warp}. The maximum limits for affine parameters are, translation $=0.75$ ($75\%$ pixel shift), scaling factor $=4$ (scaled up to $4$ times or scaled down to $1/4$\textsuperscript{th}), and shear and rotation angles are limited to $+/- \pi/3$.
 
\textbf{\textit{BO Parameters}}: {\name}'s optimization is performed for $B=50$ iterations. 
$N_0 = 10$ initial Sobol samples to start the optimization. We use Q-NoisyNegativeLogLikelihoodExpectedImprovement (QNoisyNEI), number of restarts $=50$, and raw samples $=4096$ for finding $x_{\text{EI}}$.  
In DBS ACF, a spectral gap threshold of $\gamma = 2.0$ is chosen to identify subspace dimension $d$ (Eqn. \ref{eqn:spectral_gap}). 

We use $K=4$ preference choice set and ran {\name} for three different seeded trials per prompt/target pair. 
We ran all experiments on a single NVIDIA A6000 GPU (48GB).

The {\name} image editing pipeline code base and simple inference script is released\footnote{\url{https://github.com/AnnonAnom125/annonAnomrepo.git}}.

\begin{table*}[tbp]
\centering
\caption{Quantitative comparison of different methods on the personalized image generation task (\textbf{Top}, \textcolor{blue}{second best}, \textcolor{red}{third best}).}
\small
\begin{tabular}{@{}lp{0.5cm}p{0.5cm}ccccccc@{}}
\toprule
              & \multicolumn{2}{c}{Properties} & \multicolumn{2}{c}{Target Alignment} & \multicolumn{4}{c}{Reward Metrics} & \multicolumn{1}{c}{Image Quality} \\
              \cmidrule(lr){2-3} \cmidrule(lr){4-5} \cmidrule(lr){6-9} \cmidrule(lr){10-10}
              & training-free & model-agnostic & CLIP-I ($\uparrow$) & LPIPS ($\downarrow$) & AES ($\uparrow$) & Picscore ($\uparrow$) & HPSv2 ($\uparrow$) & ImageReward ($\uparrow$) & VILA ($\uparrow$) \\ \midrule
L2-guided (Oracle)                          & \checkmarkgreen & \checkmarkgreen & 0.9811 & 0.1958 & 6.2163 & 0.2201 & 0.2646 & 0.5883 & 0.6158 \\ \midrule
IterComp                              & \crossred       & \crossred       & 0.8539 & 0.6988 & 6.1879 & 0.2289 & 0.2712 & \textcolor{blue}{1.1835} & 0.6708 \\
DiffusionDPO                          & \crossred       & \crossred       & 0.8365 & 0.7535 & 6.2367 & 0.2245 & 0.2640 & 0.6029 & 0.6560 \\ \midrule
DEMON$_{\text{Aesthetic}}$            & \checkmarkgreen & \checkmarkgreen & 0.8109 & 0.6785 & \textcolor{red}{7.2685} & 0.2224 & 0.2644 & 0.4902 & 0.6484 \\
DNO$_{\text{Aesthetic}}$              & \checkmarkgreen & \checkmarkgreen & 0.6931 & 0.8487 & \textbf{7.7674} & 0.1905 & 0.2517 & -0.9815 & 0.4509 \\
DAS$_{\text{Aesthetic}}$              & \checkmarkgreen & \checkmarkgreen & 0.6849 & 0.8495 & \textcolor{blue}{7.7032} & 0.1899 & 0.2498 & -1.2349 & 0.4514 \\
$\text{{\name}}_{\text{Aesthetic}}$   & \checkmarkgreen & \checkmarkgreen & 0.8839 & 0.6418 & 7.2313 & 0.2212 & 0.2634 & 0.7299 & 0.6649 \\ \midrule
DEMON$_{\text{PicScore}}$             & \checkmarkgreen & \checkmarkgreen & 0.8839 & 0.6742 & 6.7685 & 0.2221 & 0.2637 & 0.4805 & \textcolor{blue}{0.6792} \\
DNO$_{\text{PicScore}}$               & \checkmarkgreen & \checkmarkgreen & 0.8828 & 0.6449 & 6.2013 & \textbf{0.2326} & \textbf{0.2731} & 0.9812 & 0.6460 \\
DAS$_{\text{PicScore}}$               & \checkmarkgreen & \checkmarkgreen & 0.8758 & 0.6633 & 6.2190 & \textcolor{blue}{0.2321} & \textcolor{blue}{0.2724} & 0.9553 & 0.6425 \\
$\text{{\name}}_{\text{PicScore}}$    & \checkmarkgreen & \checkmarkgreen & 0.8877 & 0.6411 & 6.2282 & 0.2286 & 0.2679 & 0.8856 & 0.6752 \\ \midrule
DEMON$_{\text{HPSv2}}$                & \checkmarkgreen & \checkmarkgreen & 0.8472 & 0.6795 & 7.2777 & 0.2218 & 0.2627 & 0.4922 & 0.6721 \\
DAS$_{\text{HPSv2}}$                  & \checkmarkgreen & \checkmarkgreen & 0.8298 & 0.7436 & 5.9925 & 0.2128 & 0.2673 & 0.6852 & 0.6070 \\
$\text{{\name}}_{\text{HPSv2}}$       & \checkmarkgreen & \checkmarkgreen & 0.8818 & 0.6474 & 7.1735 & 0.2230 & \textcolor{red}{0.2722} & 0.8695 & \textbf{0.6806} \\ \midrule
DEMON$_{\text{ImageReward}}$          & \checkmarkgreen & \checkmarkgreen & 0.8520 & 0.6824 & 6.7221 & 0.2232 & 0.2641 & \textcolor{red}{1.1530} & 0.6642 \\
$\text{{\name}}_{\text{ImageReward}}$ & \checkmarkgreen & \checkmarkgreen & 0.8794 & 0.6468 & 6.7310 & 0.2221 & 0.2664 & \textbf{1.2275} & 0.6655 \\ \midrule
DEMON$_{\text{LPIPS}}$                & \checkmarkgreen & \checkmarkgreen & 0.9095 & \textcolor{blue}{0.5907} & 6.2229 & 0.2184 & 0.2623 & 0.3765 & 0.6154 \\
 $\text{{\name}}_{\text{CLIP-I2I}}$   & \checkmarkgreen & \checkmarkgreen & \textcolor{blue}{0.9246} & 0.6479 & 6.0338 & 0.2198 & 0.2639 & 0.8206 & 0.6186 \\
 $\text{{\name}}_{\text{LPIPS}}$      & \checkmarkgreen & \checkmarkgreen & \textcolor{red}{0.9114} & \textcolor{red}{0.5924} & 6.1711 & 0.2197 & 0.2623 & 0.8414 & 0.6250 \\ \midrule
$\text{{\name}}_{\text{Subspace}}$    & \checkmarkgreen & \checkmarkgreen & 0.8014 & 0.6726 & 6.1818 & 0.1926 & 0.2512 & 0.7863 & 0.6158 \\ 
{\name}(Ours)                         & \checkmarkgreen & \checkmarkgreen & \textbf{0.9364} & \textbf{0.5497} & 6.6690 & \textcolor{red}{0.2266} & 0.2640 & 0.8883 & \textcolor{red}{0.6723} \\ \bottomrule
\end{tabular}
\label{tab:bobench_comparison}
\end{table*}

\subsection{Results}

The two key performance goals of {\name}: Alignment with (1) target $x^*$, (2) general human preference metrics.

\circled{1} \textbf{Alignment with Target $x^*$}:

Figure \ref{fig:multi_iters} qualitatively tracks {\name}'s optimization progress. After $B=50$ iterations, {\name} steers $D_{\theta}$ to produce $\hat{x}^*$ well aligned with $x^*$. {\name} exhibits superior performance in target-based metrics, CLIP-I2I and LPIPS in Table \ref{tab:bobench_comparison} over all baselines except L2-guided, which is the upper-bound on how close we can get to the target (except DEMON$_{\text{LPIPS}}$). Preference Alignment baselines were not tuned for a particular target so their poor CLIP-I2I and LPIPS performance is expected. However, the more interesting result is that {\name} also beats {\name}$_{\text{CLIP-I2I}}$ and {\name}$_{\text{LPIPS}}$ ({\name}'s human scorer replaced by the corresponding metric scorer and directly taking target as input). This validates our hypothesis that humans function as sophisticated guidance models, capturing nuances missed by approximate metrics. Notably, users dynamically shift focus between global semantics and precise local spatial control through their preference choices.

Additionally, {\name} significantly outperforms {\name}$_{\text{Subspace}}$ on all metrics, implying that constraining the optimization to attention space instead of latent space $x_t$ is prudent to optimize quickly under $B=50$ queries for human-in-the-loop setting. (Please refer to the Appendix \ref{sec:appendix_qual_res} for qualitative results).
Thus, the versatility of human-in-the-loop multi-choice preference coupled with constrained Bayesian Optimization 
is well suited to address our target alignment task.

\circled{2} \textbf{Alignment with Preference Reward Metrics}:

 \emph{Training-based methods}: Table \ref{tab:bobench_comparison} demonstrates {\name}'s comparable performance across most alignment reward metrics while substantially outperforming fine-tuned models like DiffusionDPO \cite{wallace2024dpo}  and IterComp \cite{zhang2024itercomp} on ImageReward and Aesthetic metrics, respectively. This is particularly significant considering that DiffusionDPO and IterComp (together with a reward model) are trained on massive datasets of $58,000+$ and $55,000+$ image pairs--whereas {\name} reaches these results in only $B=50$ iterations with a single user and no model training. These results confirm that reward models remain mere proxies for human judgment; by involving humans directly, we not only improve derived metrics but also more effectively bridge the gap to true target alignment.

\emph{Inference-time methods}: 
Inference-time methods DNO \cite{tang2024inference}, DEMON \cite{yeh2024training}, and DAS \cite{kim2025test} maximize popular reward metrics rather than pursuing target alignment. While DNO and DEMON optimize diffusion noise and DAS employs Sequential Monte Carlo (SMC) to sample aligned distributions, {\name} uses probabilistic optimization. Unlike SMC's focus on distribution estimation, our BO framework directly identifies the optimum $\hat{x}^*$, making it uniquely suited for the target alignment task.

\begin{figure*}
    \centering
    \includegraphics[width=1\linewidth]{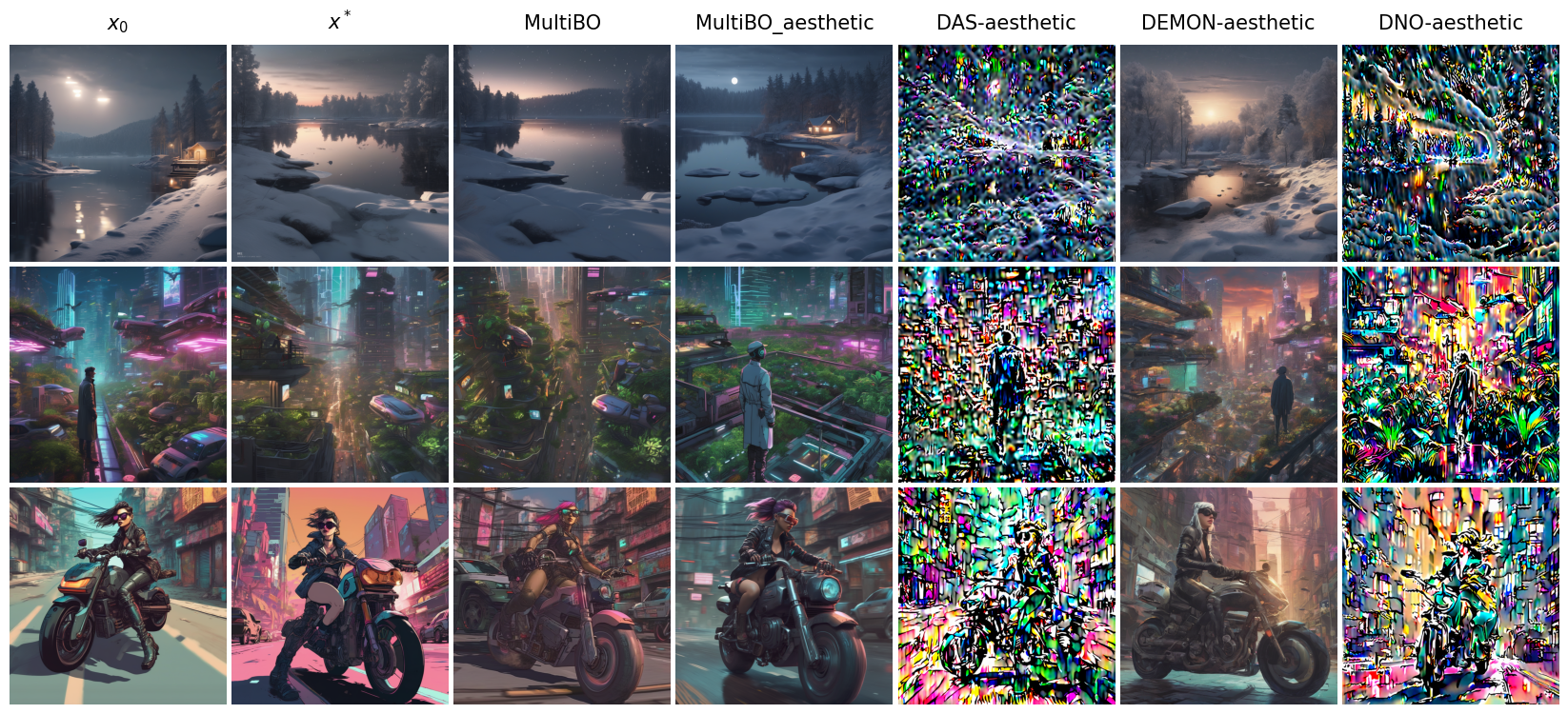}
    \caption{Qualitative comparison of {\name} ($B=50$), {\name}$_{\text{Aesthetic}}$, {DNO}$_{\text{Aesthetic}}$, {DEMON}$_{\text{Aesthetic}}$, and DNO$_{\text{Aesthetic}}$. For prompts: \textit{``A vividly realistic depiction of a snowy Swedish lake at night with hyper-detailed, cinematic-level artistry showcased on ArtStation.", ``On the rooftop of a skyscraper in a bustling cyberpunk city, a figure in a trench coat and neon-lit visor stands amidst a garden of bio-luminescent plants, overlooking the maze of flying cars and towering holograms. Robotic birds flit among the foliage, digital billboards flash advertisements in the distance.", ``A cyberpunk woman on a motorbike drives away down a street while wearing sunglasses."}}
    \label{fig:methods_aesthetic}
    \vspace{-1.7em}
\end{figure*}

For a fair comparison, alongside {\name}, we evaluate \{method\}$_{<\text{reward}>}$ pairs against {\name}$_{<\text{reward}>}$ (where the human is replaced by the corresponding $<\text{reward}>$ model in the BO loop). The results in Table \ref{tab:bobench_comparison} and Figure \ref{fig:methods_aesthetic} reveal three critical insights:\\
    \textbf{a) Broad Applicability across Rewards:} Unsurprisingly, for any given reward, {\name}$_{<\text{reward}>}$—along with DNO, DEMON, and DAS operating on that same reward—achieves peak performance for that specific metric (Table \ref{tab:bobench_comparison}). {\name}$_{<\text{reward}>}$ shows consistent cross-reward performance, demonstrating that our optimization framework is broadly applicable and comparable to existing alignment methods in maximizing diverse objectives.\\
    \textbf{b) Robustness to Reward Hacking:} Existing alignment methods are often prone to reward-hacking. As shown in Figure \ref{fig:methods_aesthetic}, DNO$_{\text{Aesthetic}}$ and DAS$_{\text{Aesthetic}}$ generate very poor images despite achieving high Aesthetic scores ($\sim7.7$) in Table \ref{tab:bobench_comparison}. In contrast, {\name} avoids this pitfall, maintaining visual integrity where others fail.\\
    \textbf{c) Efficiency of Human Feedback:} The most critical result is that {\name} (with a human scorer), despite not explicitly optimizing for any specific proxy reward, performs comparably to DNO, DAS, and DEMON on their own target metrics. This emphasizes that high-quality user-in-the-loop preference input is superior to proxy metrics. By accounting for user burden as a key design constraint, {\name} successfully extracts this high-value information to achieve robust alignment while remaining practically viable.

\begin{figure*}[hbt]
    \centering
    \includegraphics[width=1.7\columnwidth]{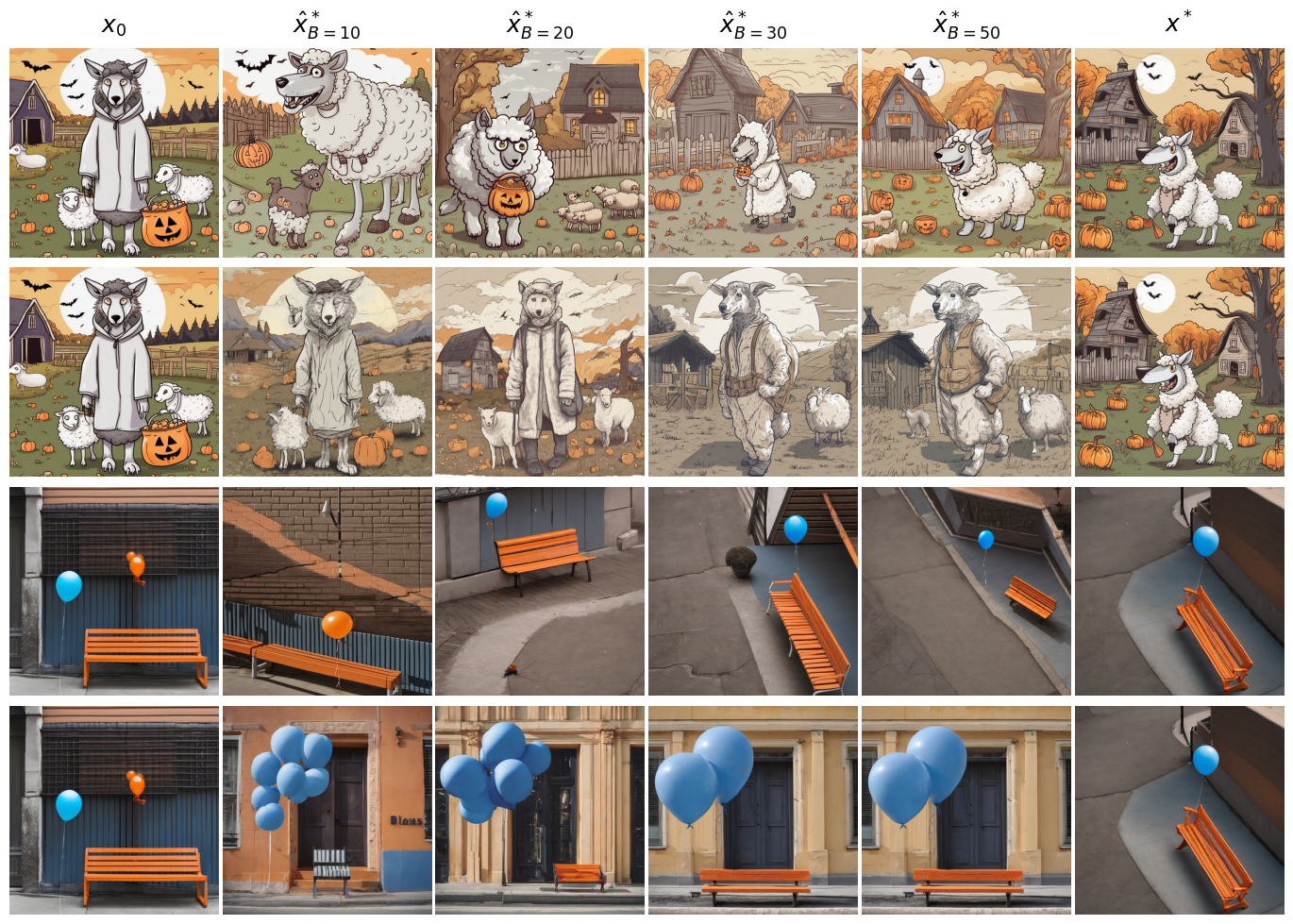}
    \caption{Qualitative results comparing {\name} (\textit{1st \& 3rd row}) and DEMON \textit{choose generate}(\textit{2nd \& 4th row}). For prompts: \textit{``A wolf wearing a sheep halloween costume going trick-or-treating at the farm", ``a blue balloon and a orange bench".}}
    \label{fig:methods_demon}
\end{figure*}

For a true analysis of {\name}'s target and reward alignment, we compare against DEMON's \textit{choose generate} mode, where user selection acts as a non-differentiable reward ($+1/-1$), mirroring our PairwiseGPR formulation (Eqn. \ref{eqn:pbo-likelihood}). Figure \ref{fig:methods_demon} shows that {\name} converges closer to $x^*$ within $50$ iterations, whereas DEMON stalls around iteration $30$. This confirms that naive human-in-the-loop integration is insufficient; effective optimization must balance editing freedom with user burden, which {\name} achieves via MultiwiseGPR and DBS ACF. Furthermore, unlike DEMON's high-dimensional latent noise ($\epsilon_t$) optimization, {\name} operates in a lower-dimensional attention space, enabling significantly faster alignment.

$\blacksquare$ \textbf{Human Evaluation}:  

We report human evaluation results from $30$ volunteers. Participants were asked to pick the Top-2 methods closest to the target $x^*$. A method is ranked 1, 2 or 3 (not in top 2). Table \ref{tab:human_eval} demonstrates that {\name} has a high win rate compared to baselines with high user agreement.

Please refer to the Appendix \ref{sec:appendix_qual_res} for additional results including qualitative results on all other reward metrics.

\begin{table}[hbt]
\centering
\caption{Human evaluation results: We report the Win Rate, Mean Rank, and MOS. Kendall’s $W = 0.65$ indicates strong inter-rater agreement.}
\label{tab:human_eval}
\small
\begin{tabular}{lccc}
        \toprule
        \textbf{Method} & \textbf{Win Rate (\%)} $\uparrow$ & \textbf{Mean Rank} $\downarrow$  & \textbf{MOS} $\uparrow$ \\ 
        \midrule
                
        DAS$_{\text{PicScore}}$ & 0.53 & 2.98  & 1.03 \\
        DNO$_{\text{PicScore}}$ & 5.36 & 2.87  & 1.25 \\
        IterComp & 6.49 & 2.81  & 1.38 \\
        DEMON$_{\text{Aesthetic}}$ & 9.55 & 2.79  & 1.42 \\
        {\name}$_{\text{Aesthetic}}$ & 11.26 & 2.73 & 1.53 \\
        DEMON$_{\text{choose}}$ & 15.60 & 2.67  & 1.66 \\
        {\name}$_{\text{LPIPS}}$ & 31.81 & 2.36  & 2.27 \\
        {\name} (Ours) & \textbf{70.82} & \textbf{1.31}  & \textbf{3.58} \\
        \bottomrule
\end{tabular}
\end{table}

 \begin{figure}[tbp]
    \centering
    \includegraphics[width=1\linewidth]{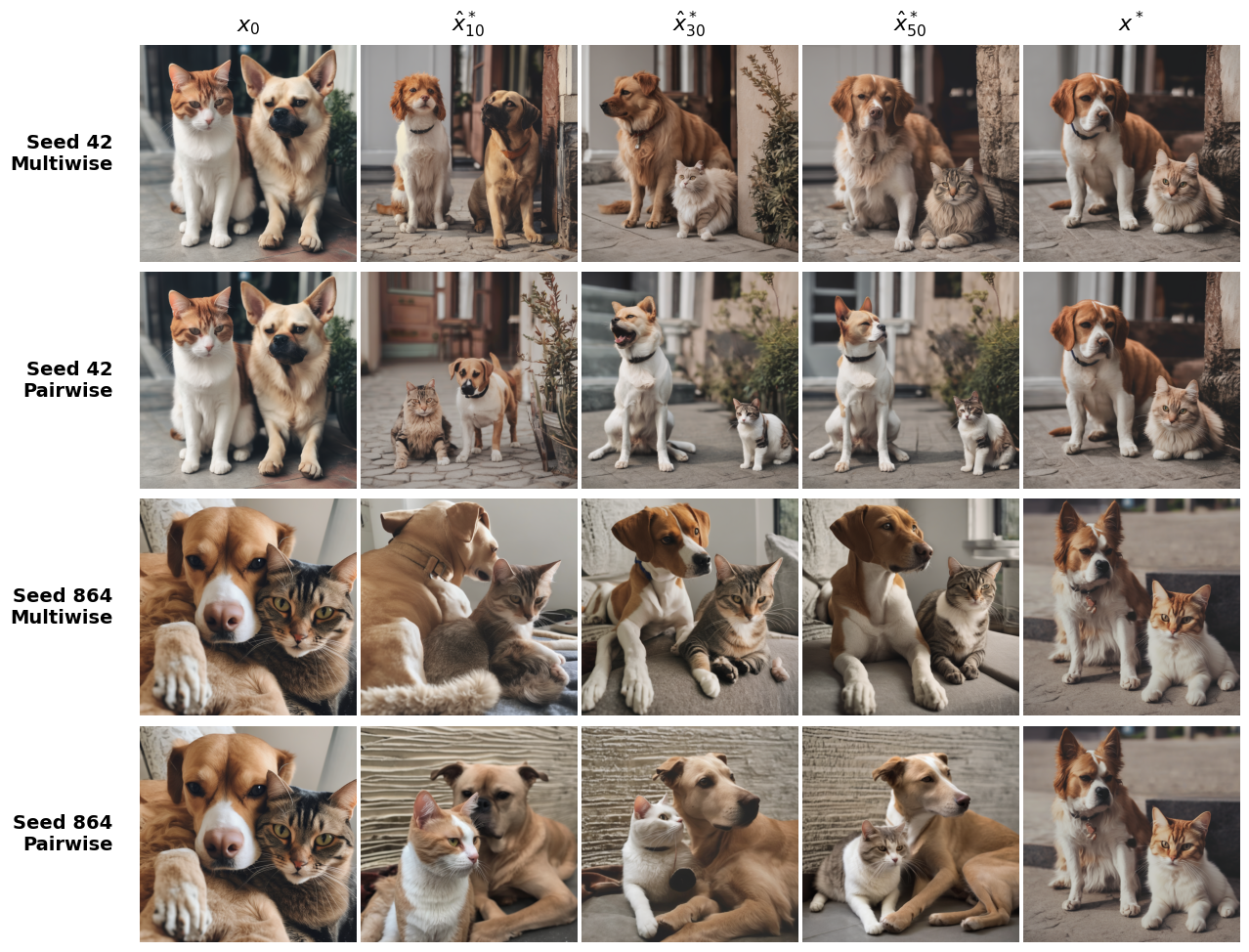}
    \caption{Ablation - Pairwise vs Multiwise Optimization Progress for two different seeds}
    \label{fig:ablation_gpr}
\end{figure}

$\blacksquare$ \textbf{Ablation Studies}:

(1) \emph{PairwiseGPR likelihood ($K=2$) vs MultiwiseGPR ($K>2$)}: 
\label{sec:ablation}
In $B=50$ steps, Figure \ref{fig:ablation_gpr} and and Table \ref{tab:ablation_gpr} illustrate that PairwiseGPR requires far more preference feedback, $B$ to sufficiently span the optimization input space for accurately modeling $f$ and identifying it's optimum $x^*$. 

\begin{table}[ht]
    \centering
    \small
    \begin{minipage}{0.45\columnwidth}
        \centering
        \begin{tabular}{|c|c|c|} 
            \hline
            Method & CLIP-I & LPIPS \\ \hline \hline
            Pairwise & 0.81 & 0.68 \\ \hline
            Multiwise & \textbf{0.94} & \textbf{0.55} \\ \hline
        \end{tabular}
        \caption{PairwiseGPR vs MultiwiseGPR performance on target-based metrics}
        \label{tab:ablation_gpr}
    \end{minipage}
    \hfill
    \begin{minipage}{0.45\columnwidth}
        \centering
        \begin{tabular}{|c|c|}
            \hline
            K & LPIPS \\ \hline \hline
            4 & 0.55 \\ \hline
            6 & 0.54 \\ \hline
            10 & 0.64 \\ \hline
        \end{tabular}
        \caption{{\name} performance for $K$ preference choices}
        \label{tab:choice-hyp}
    \end{minipage}
\end{table}


(2) \emph{Number of choices $K$ in choice set $\mathcal{Z}$}: As number of choices increases, the reliability of user preference drops as they have too much choice, this is reflected in poor metrics reported in Table \ref{tab:choice-hyp} for $K=10$. $K=6$ only fetches marginal gains disproportionate to user burden. $K=4$ offers good trade-off between informative preference data and minimal user burden.

\begin{figure}[H]
    \centering
    \caption{Ablation - {\name} applied to different attention layers two different seeds. ($t^+ = 0.0, \Delta=0.2$, MultiwiseGPR. For prompt: \textit{``a cat and a dog"})}
    \includegraphics[width=1\linewidth]{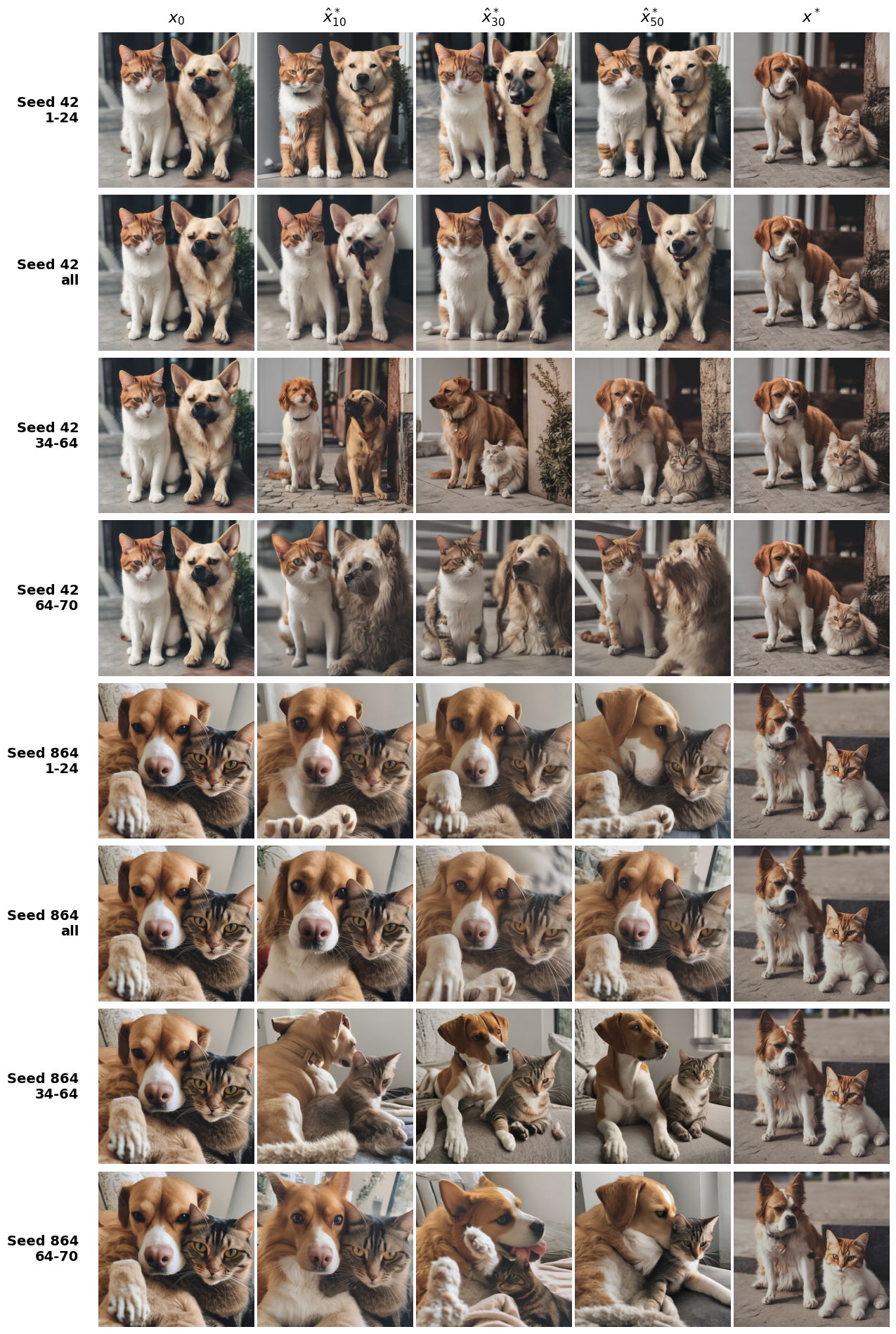}
    
    \label{fig:ablation_attn_layers}
    \vspace{-2em}
\end{figure}

(3) \emph{Self-Attention Layers to Edit}: Middle to later self-attention layers (decoder layers of UNet) are ideal for spatial and semantic control. The early attention layers do not hold enough semantic structure to cause precise change, often resulting in complete loss of similarity to the source image while the very last layers only result in minute peripheral changes. Figure \ref{fig:ablation_attn_layers} shows the complete loss of semantic structure when $1-24$ attention layers of SDXL model are modified and the lack of significant changes when editing layers $64-70$. Layers $34-64$ offer the most editing advantage.

\begin{figure}
    \centering
    \includegraphics[width=1\linewidth]{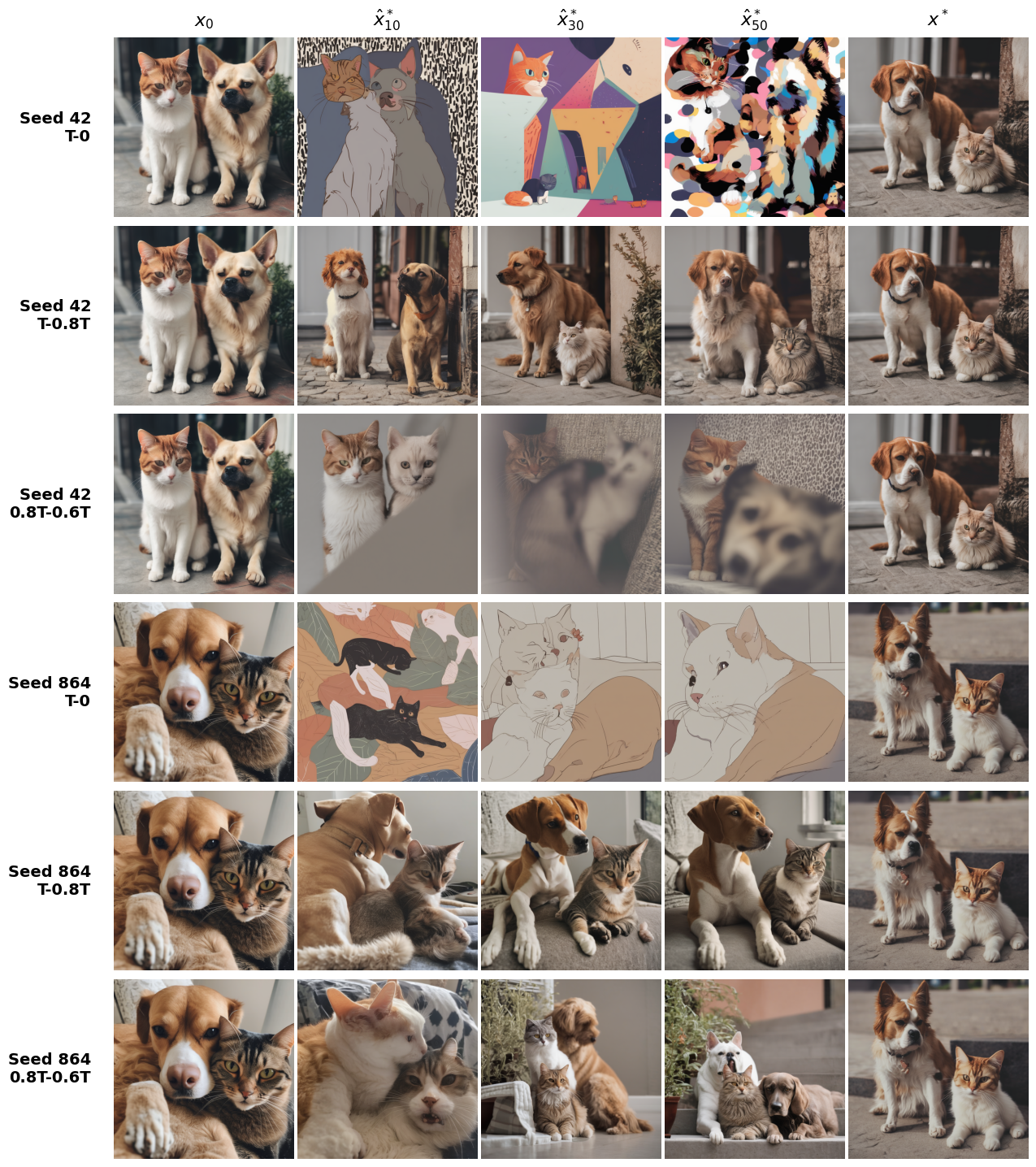}
    \caption{Ablation - {\name} applied at different timesteps $t$ and intervals $\Delta$. (Attn layers $34-64$, MultiwiseGPR)}
    \label{fig:ablation_time}
\end{figure}

(4) \emph{Timesteps to Edit}: Figure \ref{fig:ablation_time} identifies that the earlier ($T$ to $0.8T$) we perform edits the more freedom we have in steering the diffusion generation. Editing attention should not be done for prolonged number of timesteps ($T$ to 0) as it steers the diffusion process to low probability points. 

%% file: tex/_5_related_works.tex
\section{Related Works}
\label{sec:related-work}

$\blacksquare$ \textbf{RLHF and Reward Based Methods}: RLHF \cite{bai2022training,ouyang2022training} has been extensively adapted for T2I diffusion \cite{domingo2025adjoint,jiang2024realigndiff,li2024reward,uehara2024feedback,zhang2024aligning,lee2023aligning,sun2023dreamsync}. Methods like ImageReward \cite{xu2023imagereward,clark2023direct,prabhudesai2023rewardbackprop} use supervised loss with trained reward models. Others like DDPO \cite{black2023trainingdiffusion} and DPOK \cite{fan2023dpok,fan2024rlfinetune} formulate sampling as a Markov decision process, applying RL to maximize rewards.

$\blacksquare$ \textbf{Preference Alignment in Diffusion Models}: Aligning diffusion with human preferences is critical for enhancing generation quality and user satisfaction. Direct Preference Optimization (DPO) methods like DiffusionDPO \cite{wallace2024dpo} bypass reward model training, directly finetuning on preference data. IterComp \cite{zhang2024itercomp} and CaPO \cite{lee2025calibrated} extend this by iteratively training rewards or calibrating preferences without annotations. To mitigate reward over-optimization, recent works use sample inversion techniques: DDIM-InPO \cite{lu2025inpo}, SmPO-Diffusion \cite{lu2025smoothed}, InversionDPO \cite{li2025inversion} reformulate DPO loss using inversion or smoothed distributions. Training-free approaches target inference time. DAS \cite{kim2025test} uses Sequential Monte Carlo, while DNO \cite{tang2024inference} optimizes noise via reward guidance. DEMON \cite{yeh2024training} proposes stochastic optimization to guide denoising without backpropagation. However, challenges regarding reward hacking, computational overhead, and personalization persist.

\noindent $\blacksquare$ \textbf{Training-free Attention based Methods}: Attention-based spatial editing employs strategies like feature injection \cite{zhou2025attention,chen2023fec,cao2023masactrl,huang2023kvinversion,khandelwal2023infusion,wu2024gaussctrl,long2024wonder3d,tumanyan2023plugplay}, attention map operations \cite{wu2024gaussctrl,mou2024diffeditor}, feature concatenation \cite{balaji2022ediffi, mou2023dragondiffusion, deng2023zstar}, cross-modal optimization \cite{hertz2022prompttoprompt,esser2024scaling}, and adapter-based re-learning \cite{ye2024ipadapter}.
Works such as DreamMatcher \cite{nam2024dreammatcher} achieves semantic alignment via source $Q,K$ injection and $V$ warping. While effective at reducing inconsistencies, such methods lack free-form editing capabilities, typically relying on source images or custom instructions.


$\blacksquare$ \textbf{Guidance-based Methods}:
Many guidance methods build on the score-based formulation of diffusion models  \cite{song2021score}. Classifier Guidance \cite{dhariwal2021diffusion} requires additional training, works like Universal Guided Diffusion \cite{he2024manifold,yoon2023censored,song2023lossguided,yu2023freedom,chung2023posterior} approximate guidance to use off-the-shelf classifiers or models directly. These methods use Tweedie’s formula operate on the predicted clean sample
However, such methods suffer from poor guidance, and the Tweedie mean prediction quality limits effectiveness in maximizing complex rewards.
\textit{\textbf{Training-free Spatial Editing methods:}}InstanceDiffusion ~\citep{instance-diffusion} uses instance-level anchors for specialized object placement and scene construction. Works like ~\citep{box-diff,attention-refocus,dense-diffusion,loco-layout} use specialized conditioning modalities like bounding boxes, vectors, masks, box-text pairs, custom instructions, etc. to explicitly steer the diffusion generation. These methods place coonsiderable burden on the user to provide these inputs and often perform poorly when a combination of changes are required or when the user is unable to express their requirement. Works like ~\citep{rpg,ella}
~\citep{rpg}; ~\citep{layoutllm}, and \cite{yu2025anyedit} use Mixture of Experts (MoEs) or LLMs to  parse the user requirement into several sub-editing tasks assigned to a specific editing pipeline that excels in that task. Although these methods offer an interactive editing framework they are often limited by the user's capability to express the required changes via syntax like language. 

%% file: tex/conclusion.tex
%

\section{Conclusions}
We propose {\name}, a training-free, human-in-the-loop image personalization framework built upon Preferential Bayesian Optimization. {\name} addresses the “last-mile” gap between a user’s latent visual intent and the sub-optimal images produced by prompt-based generative models. With only 50 preference queries, the method converges to images that closely align with the user’s ideal target, without requiring any task-specific training data. Query efficiency is further improved by incorporating multi-choice preference feedback and constraining the attention optimization to space of valid warping transformations, substantially reducing the search complexity.

\section{Limitations and Future Work}
While {\name} achieves good target alignment within a budget $B$, Figure \ref{fig:failure} highlights room for improvement. We diagnose the problem as twofold. First, there are harder scenarios where the generation is reasonably close to the target and the user is forced to choose between equally bad or good choices, i.e., the preference signal weakens, stalling the optimization. Secondly, more iterations $B>50$ may be needed. This is influenced by the decision to optimize only in the attention domain and restricting to a transformation space, limiting the arsenal of possible changes that can be affected.

For future work, we plan to address these issues by incorporating informed priors into the Gaussian likelihood of BO by leveraging pre-trained reward models. In addition, we aim to explore customized high-dimensional Bayesian optimization strategies that are better suited to the structure and geometry of latent diffusion representations.

%% file: tex/appendix.tex
\section{Appendix}
\subsection{Gaussian Process Likelihoods \& Posterior}
We derive the GPR likelihood expressions for MultiwiseGPR modeling 1-out-of-$K$ preference choices in \ref{sec:appendix_multiwise}, and SubsetMultiwiseGPR modeling $N$-out-of-$K$ preference choices in \ref{sec:appendix_subset} respectively.

\subsubsection{Pairwise Binomial Logit Likelihood GPR}
\label{sec:appendix_pairwise}
The GPR posterior \ref{eqn:pbo-posterior} constructed from the binomial likelihood in Eqn \ref{eqn:pbo-likelihood} has no closed-form expression and although there exist sophisticated variational and Monte Carlo methods, it is typically estimated by Laplace approximation \cite{brochu2010tutorial}.  The Laplace approximation follows from Taylor-expansion of the log-posterior of GPR about a set point $\hat{\mathbf{f}} = \mathbf{f}_{\text{MAP}}$, the MAP estimation and is given as,
\begin{equation}
    \begin{aligned}
        \log P(\mathbf{f}|\mathcal{X}) &= \log P(\hat{\mathbf{f}} | \mathcal{X}) + \mathbf{g}^T(\mathbf{f}-\hat{\mathbf{f}})\\ &- \frac{1}{2}(\mathbf{f}-\hat{\mathbf{f}})^T\mathbf{H}(\mathbf{f}-\hat{\mathbf{f}})
    \end{aligned}
    \label{eqn:laplace-approx}
\end{equation}

where, $\mathbf{g}, \mathbf{H}$ are the gradient and Hessian respectively. Their closed form expressions are found in \cite{chu2005preference},


\subsubsection{Multiwise Logit Likelihood GPR}
\label{sec:appendix_multiwise}
We derive the likelihood expression in Eqn. \ref{eqn:multi-likelihood} as follows,

Consider a choice set $\mathcal{Z} = \{1,\dots,K\}$.
Each choice $j \in \mathcal{Z}$ is associated with a latent utility
\begin{equation}
v_j = f_j + \delta_j,
\end{equation}
where $f_j \in \mathbb{R}$ is a true function value and $\delta_j$ is random noise.
The observed choice is
\begin{equation}
a = \arg\max_{j \in \mathcal{Z}} v_j.
\end{equation}

The probability that choice $i$ is chosen is
\begin{equation}
\begin{aligned}
P(a = i \mid \mathbf f)
&= P(v_i \ge v_j \;\; \forall j \neq i) \\
&= P(f_i + \delta_i \ge f_j + \delta_j \;\; \forall j \neq i) \\
&= P(\delta_j \le \delta_i + f_i - f_j \;\; \forall j \neq i).
\end{aligned}
\end{equation}

\begin{equation}
\begin{aligned}
P(a = i \mid \mathbf f)
&=
\int
P(\delta_j \le \delta_i + f_i - f_j \;\; \forall j \neq i \mid \delta_i)
\, p(\delta_i) \, d\delta_i \\
&=
\int
\prod_{j \in \mathcal{Z},\, j \neq i}
F(\delta_i + f_i - f_j)
\; p(\delta_i) \, d\delta_i,
\end{aligned}
\end{equation}
where $F(\cdot)$ is the CDF of $\delta_j$.

Assume that $\delta_j \sim \text{IID Gumbel}(0,1)$, with CDF and PDF,
\begin{equation}
F(\delta) = \exp(-e^{-\delta}), 
\qquad
p(\delta) = \exp\big(-(\delta + e^{-\delta})\big).
\end{equation}

Substituting them,
\begin{equation}
\begin{aligned}
P(a = i \mid \mathbf f)
=
\int
&\prod_{j \in \mathcal{Z},\, j \neq i}
\exp\!\left(-e^{-(\delta_i + f_i - f_j)}\right)
\\
&\exp\!\left(-(\delta_i + e^{-\delta_i})\right)
\, d\delta_i.
\end{aligned}
\end{equation}

Simplifying,
\begin{equation}
\prod_{j \in \mathcal{Z},\, j \neq i}
\exp\!\left(-e^{-(\delta_i + f_i - f_j)}\right)
=
\exp\!\left(
- e^{-\delta_i} e^{-f_i} \sum_{j \in \mathcal{Z},\, j \neq i} e^{f_j}
\right).
\end{equation}

Thus,
\begin{equation}
\begin{aligned}
P(a = i \mid \mathbf f)
&=
\int
\exp\!\left(
-\delta_i
-
e^{-\delta_i}
\Big(1 + e^{-f_i} \sum_{j \in \mathcal{Z},\, j \neq i} e^{f_j}\Big)
\right)
\, d\delta_i.
\end{aligned}
\end{equation}

Change of variables $t = e^{-\delta_i}$, and $d\delta_i = -dt/t$.
The integral becomes,
\begin{equation}
\begin{aligned}
P(a = i \mid \mathbf f)
&=
\int_{0}^{\infty}
\exp\!\left(
- t \Big(1 + e^{-f_i} \sum_{j \in \mathcal{Z},\, j \neq i} e^{f_j}\Big)
\right)
\, dt.
\end{aligned}
\end{equation}

which gives,
\begin{equation}
\begin{aligned}
P(a = i \mid \mathbf f)
&=
\frac{1}{1 + e^{-f_i} \sum_{j \in \mathcal{Z},\, j \neq i} e^{f_j}}
=
\frac{e^{f_i}}{\sum_{j \in \mathcal{Z}} e^{f_j}}.
\end{aligned}
\label{eqn:multi-likelihood-derive}
\end{equation}

Therefore, the multinomial-logit probability is the softmax function (Eqn. \ref{eqn:multi-likelihood})
\begin{equation}
P(a = i \mid \mathbf f)
=
\frac{\exp(f_i)}{\sum_{j \in \mathcal{Z}} \exp(f_j)}.
\end{equation}

We use Laplace Approximation (Eqn. \ref{eqn:laplace-approx}) to estimate the posterior in Eqn. \ref{eqn:multi-posterior}.
The corresponding gradient and Hessian are,

\begin{equation}
\label{eq:posterior-gradient}
\nabla \log p(\mathbf f \mid \mathcal X) = - \mathbf{K}^{-1} \mathbf{f} + \sum_{i=1}^{M} \left( \mathbf e_{a_i} -\boldsymbol{\pi}_i \right),
\end{equation}

\begin{equation}
\label{eq:posterior-hessian}
\nabla^2 \log p(\mathbf{f} \mid \mathcal{X}) = - \mathbf{K}^{-1} 
- \sum_{i=1}^{M} \left[ \mathrm{diag}(\boldsymbol{\pi}_i) - \boldsymbol{\pi}_i \boldsymbol{\pi}_i^\top \right],
\end{equation}

where, $\mathbf{K}$ is the GPR prior, $\mathbf e_{a_i}$ is the one-hot indicator vector for the observed choice $a_i \in \mathcal Z_i$, i.e., the winner $a_i$ in $i$th choice set $\mathcal{Z}_i$ is $x_j$ from the set of $N$ samples $[x_1,\dots,x_N]$,
$(\mathbf e_{a_i})_j = \mathbf{1}[x_j = a_i]$.
and,
\begin{equation}
(\pi_i)_j = P(a = x_j|\mathbf f) = \frac{\exp(f_j)}{\sum_{k \in \mathcal{Z}_i} \exp(f_k)}
\end{equation}
is the softmax function.

\subsubsection{Subset-Choice Multiwise Logit Likelihood GPR}
\label{sec:appendix_subset}

We consider a more general form of multiwise preference scenario, where instead of selecting one choice, the user selects a \emph{non-empty subset} of the $K$ choices.

Let $\mathcal{Y} = \{x_1, \dots, x_N\}$ be a set of $N$ samples.
A single choice set of size $K$ is,
\[
\mathcal{Z} = \{a^{(1)}, \dots, a^{(K)}\} \subseteq \mathcal{Y}.
\]
The user preference is a non-empty subset
\[
A \subseteq \mathcal{Z}, \qquad A \neq \emptyset.
\]

The observation to the likelihood model is,
\[
\mathcal{X} = \{ A \succ \emptyset \mid A \subseteq \mathcal{Z} \}.
\]

We assume a Gaussian prior over the latent utility function $f$,
\[
f \sim \mathcal{N}(0,\mathbf{K}),
\]
The latent noisy function values are,
\[
\mathbf f = \{ f(x) : x \in \mathcal Z \}.
\]
\[
v(x) = f(x) + \delta,
\]
where the noise is IID $\delta \sim$ Gumbel$(0,1)$,

Unlike the multinomial-logit model, which assumes a single winning choice, we
assume that each element in $\mathcal{Z}$ is independently selected or rejected,
and the chosen subset $A$ corresponds to the selected elements.

For a \emph{subset} $A \subseteq \mathcal{Z}$, we define the observed choice as:
\[
x_j \in A , \quad v_j \text{ is ``chosen"}.
\]

Assuming IID Gumbel$(0,1)$ noise, the probability
of observing subset $A$ is same as the multinomial-logit derivation in Eqn. \ref{eqn:multi-likelihood-derive}. The likelihood of a sample $x_j$ to be in the chosen subset $x_j \in A$ is,


\[
P(x_j \in A \mid \mathbf f) \propto \exp(f_j),
\]
 Assuming each sample $x_j$ in choice set $\mathcal Z$ is independently selected or not, to be part of subset $A$, the joint probability of all samples in $A$ is then,
\[
\prod_{x_j \in A} \exp(f_j) = \exp\Big(\sum_{x_j \in A} f_j\Big).
\]

The likelihood probability normalization factor (similar to denominator of Eqn. \ref{eqn:multi-likelihood-derive}) is obtained by summing over all non-empty subsets of $\mathcal Z$ as $A$ is one such subset,
\[
\sum_{\emptyset \neq C \subseteq \mathcal Z} \exp\Big(\sum_{x_j \in C} f_j\Big).
\]

Simplifying,
\[
\sum_{C \subseteq \mathcal Z} \prod_{x_j \in C} e^{f_j} = \prod_{x_j \in \mathcal Z} (1 + e^{f_j}),
\]
and removing the empty set case, it becomes,
\[
\prod_{x_j \in \mathcal Z} (1 + e^{f_j}) - 1.
\]

Thus, the likelihood of observing a subset
$A \subseteq \mathcal{Z}$ is,
\begin{equation}
P(A \mid \mathbf f) 
= \frac{\exp\Big(\sum_{x_j \in A} f_j\Big)}
       {\prod_{x_j \in \mathcal Z} (1 + e^{f_j}) - 1}.
\end{equation}



This likelihood reduces to the multinomial-logit model (Eqn. \ref{eqn:multi-likelihood}) when $|A| = 1$.

Given a collection of $M$ observations
$\mathcal{X} = \{A_i, \mathcal Z_i\}_{i=1}^M$,
the posterior distribution is
\begin{equation}
\label{eqn:subset-posterior}
P(\mathbf f \mid \mathcal X)
=
\frac{P(\mathbf f)}{P(\mathcal X)}
\prod_{i=1}^M
P(A_i \mid \mathbf f_{\mathcal Z_i}),
\end{equation}
where $P(\mathbf f)$ is the GP prior and
$P(\mathcal{X}) = \int P(\mathcal{X}|\mathbf{f})P(\mathbf{f})d\mathbf{f}$.

Same as MultiwiseGPR with one choice, we use Laplace Approximation (Eqn. \ref{eqn:laplace-approx}) to estimate the posterior in Eqn. \ref{eqn:subset-posterior}.

We now compute the Gradient and Hessian of the posterior.
For a single observation $(A,\mathcal Z)$, we define the indicator vector
$\mathbf Y \in \{0,1\}^{|\mathcal Z|}$ as,
\[
Y_j =
\begin{cases}
1, & x_j \in A, \\
0, & \text{otherwise}.
\end{cases}
\]

The gradient and Hessian are,
\begin{equation}
\begin{aligned}
\nabla \log p(\mathbf f \mid \mathcal X)
=
- \mathbf K^{-1} \mathbf f
+ \sum_{i=1}^M \left( \mathbf Y_i - \boldsymbol{\pi}_i \right).
\end{aligned}
\end{equation}

\begin{equation}
\begin{aligned}
\nabla^2 \log p(\mathbf f \mid \mathcal X)
=
- \mathbf K^{-1}
- \sum_{i=1}^M \Gamma_i .
\end{aligned}
\end{equation}

where, $\pi_j$ are defined as,
\begin{equation}
\begin{aligned}
\pi_j
&=
P(x_j \in A \mid \mathbf f) \\
&=
\frac{
\sum_{\emptyset \neq C \subseteq \mathcal Z,\; C \ni x_j}
\exp\!\left( \sum_{x \in C} f(x) \right)
}{
\sum_{\emptyset \neq C \subseteq \mathcal Z}
\exp\!\left( \sum_{x \in C} f(x) \right)
}.
\end{aligned}
\end{equation}
where, $ \emptyset \neq C_i \subseteq \mathcal{Z}_i, C_i \ni x_j $ means valid non-empty subsets of $\mathcal{Z}_i$ that contain $x_j$ in them (For $M$ observations, we have $(\pi_i)_j$). 

The Hessian diagonal and off-diagonal entries are,
\begin{equation}
\begin{aligned}
\Gamma_{jj} &= \pi_j (1 - \pi_j), \\
\Gamma_{jk} &= \pi_{jk} - \pi_j \pi_k, \quad j \neq k,
\end{aligned}
\end{equation}
where,
\[
\pi_{jk} = P(x_j, x_k \in A \mid \mathbf f)
\]

\begin{figure*}[htbp]
    \centering
    \includegraphics[width=0.6\linewidth]{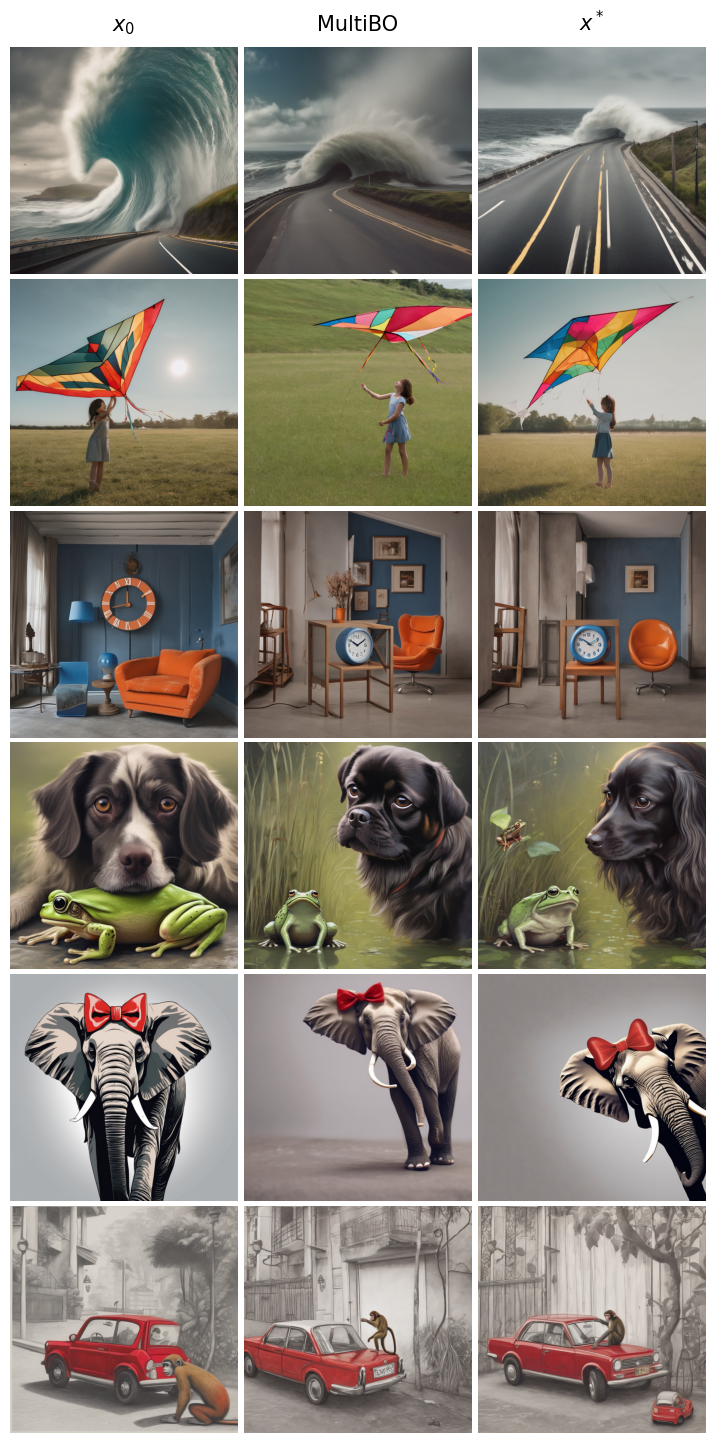}
    \caption{Qualitative results of {\name} ($B=50$). For prompts: \textit{"a tidal wave approaching a coastal road","A girl is holding a large kite on a grassy field.","a orange chair and a blue clock","a dog and a frog","a elephant and a bow","a monkey and a red car".}}
    \label{fig:appendix_multibo}
\end{figure*}

\section{More Qualitative Results}
\label{sec:appendix_qual_res}
We show qualitative results for {\name}, {\name}$_{<\text{reward}>}$, training methods like DiffusionDPO \cite{wallace2024dpo} and IterComp \cite{zhang2024itercomp} and inference time methods, DNO \cite{tang2024inference}, DAS \cite{kim2025test}, DEMON \cite{yeh2024training} operating on different ${<\text{reward}>}$ metrics. These results complement the quantitative counterparts reported in Table \ref{tab:bobench_comparison} in the main paper.
\begin{figure*}
    \centering
    \includegraphics[width=1\linewidth]{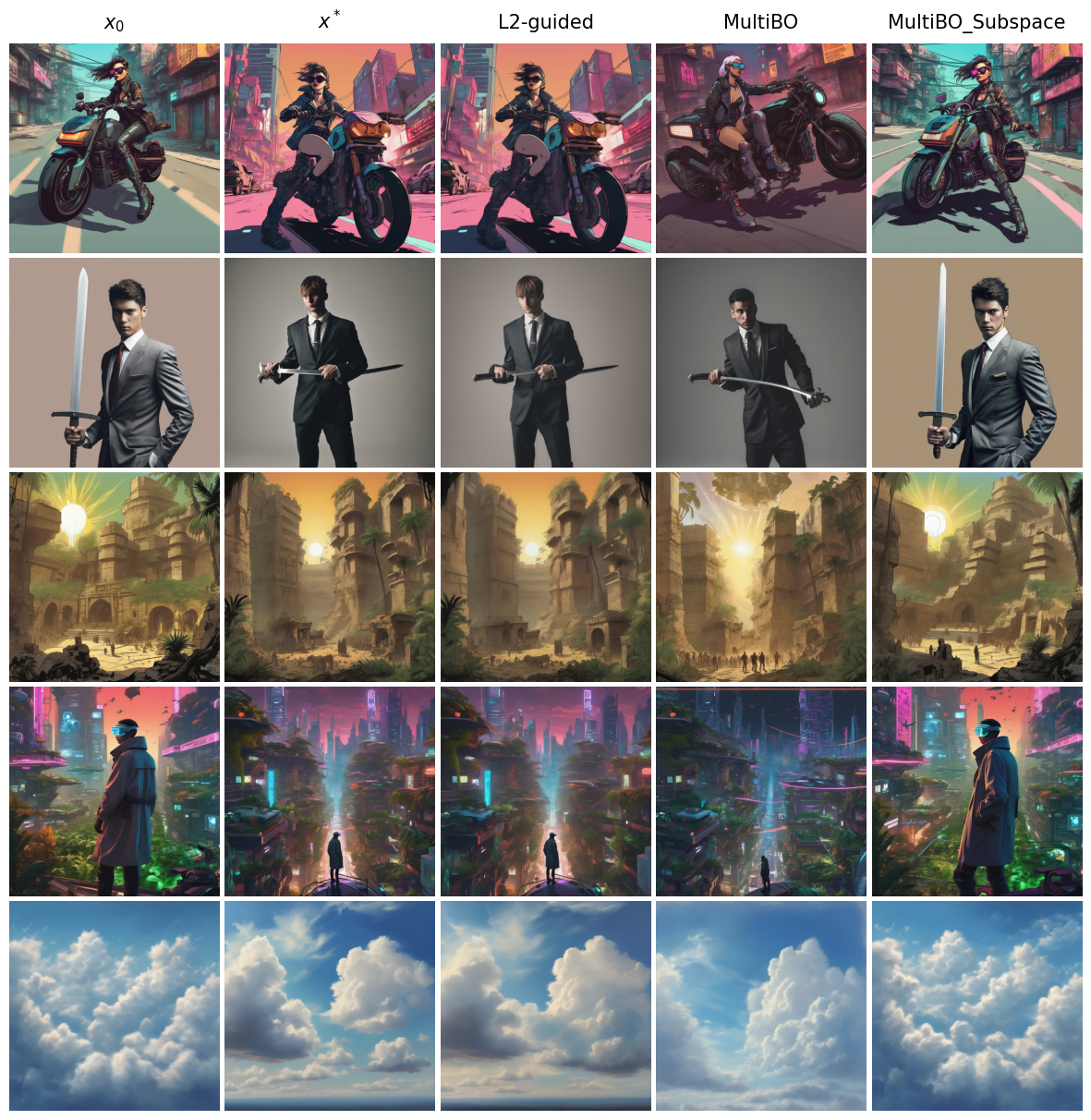}
    \caption{Qualitative results comparing {\name} ($B=50$), L2-guided, and {\name}$_{\text{Subspace}}$. For prompts: \textit{``A cyberpunk woman on a motorbike drives away down a street while wearing sunglasses.",``A person in a suit holding a sword.",``A comic book illustration by John Kirby depicting a jungle fortress surrounded by dirt walls in a marketplace setting with cinematic rays of sunlight.",``On the rooftop of a skyscraper in a bustling cyberpunk city, a figure in a trench coat and neon-lit visor stands amidst a garden of bio-luminescent plants, overlooking the maze of flying cars and towering holograms. Robotic birds flit among the foliage, digital billboards flash advertisements in the distance.",``The soft, fluffy clouds drifted lazily across the blue sky, a canvas of endless possibilities and imagination."}}
    \label{fig:appendix_l2}
\end{figure*}
\begin{figure*}
    \centering
    \includegraphics[width=1\linewidth]{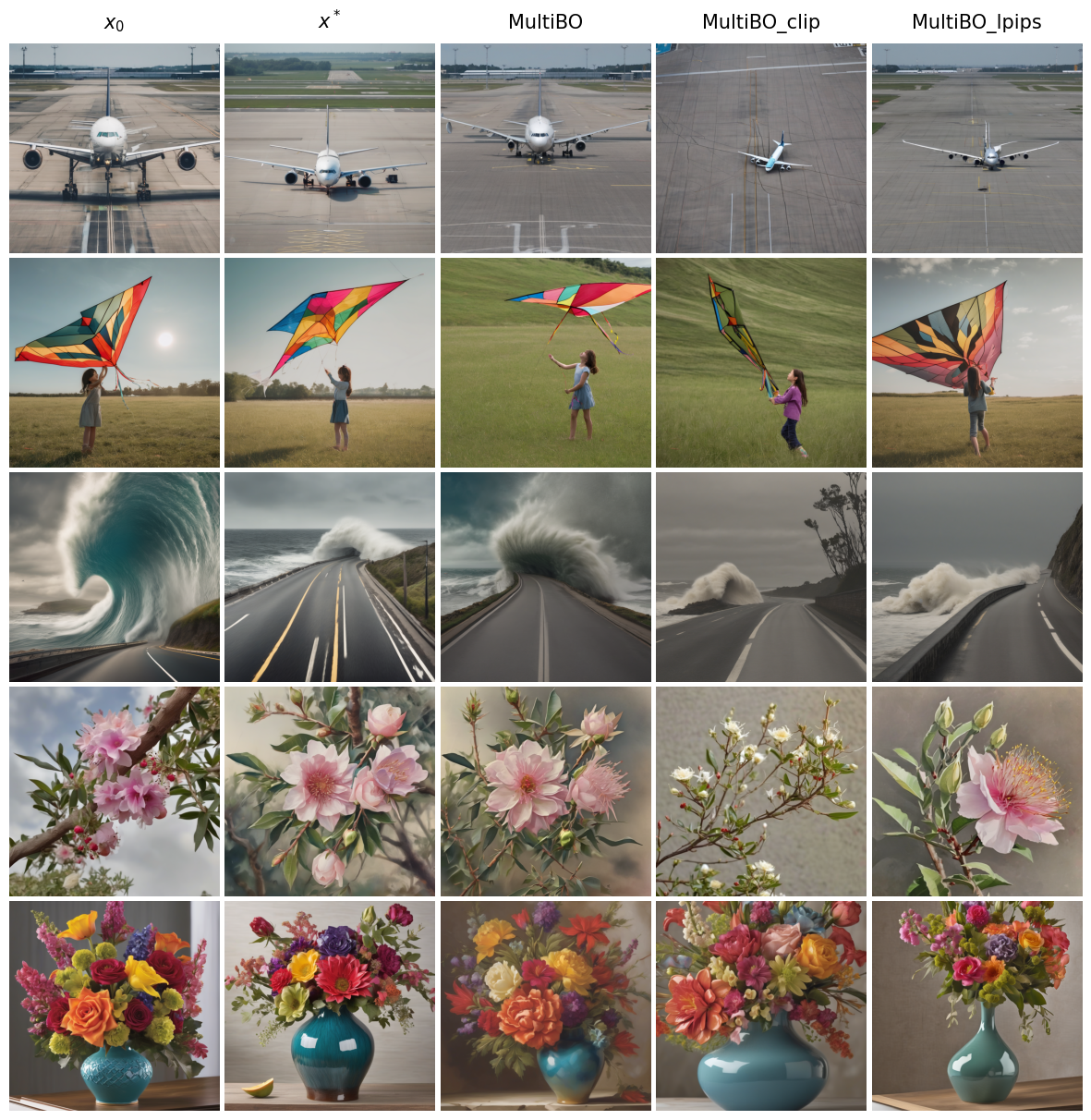}
    \caption{Qualitative results comparing {\name} ($B=50$), {\name}$_{\text{CLIP-I}}$, and {\name}$_{\text{LPIPS}}$. For prompts: \textit{``An airplane on the runway of an airport.",``A girl is holding a large kite on a grassy field.",``a tidal wave approaching a coastal road",``The fragrant flowers bloomed on the sturdy stem and the thorny bush.",``The smooth, glossy finish of the ceramic vase accentuated the vibrant colors of the flowers, a stunning centerpiece of beauty."}}
    \label{fig:appendix_lpips}
\end{figure*}
\begin{figure*}
    \centering
    \includegraphics[width=1\linewidth]{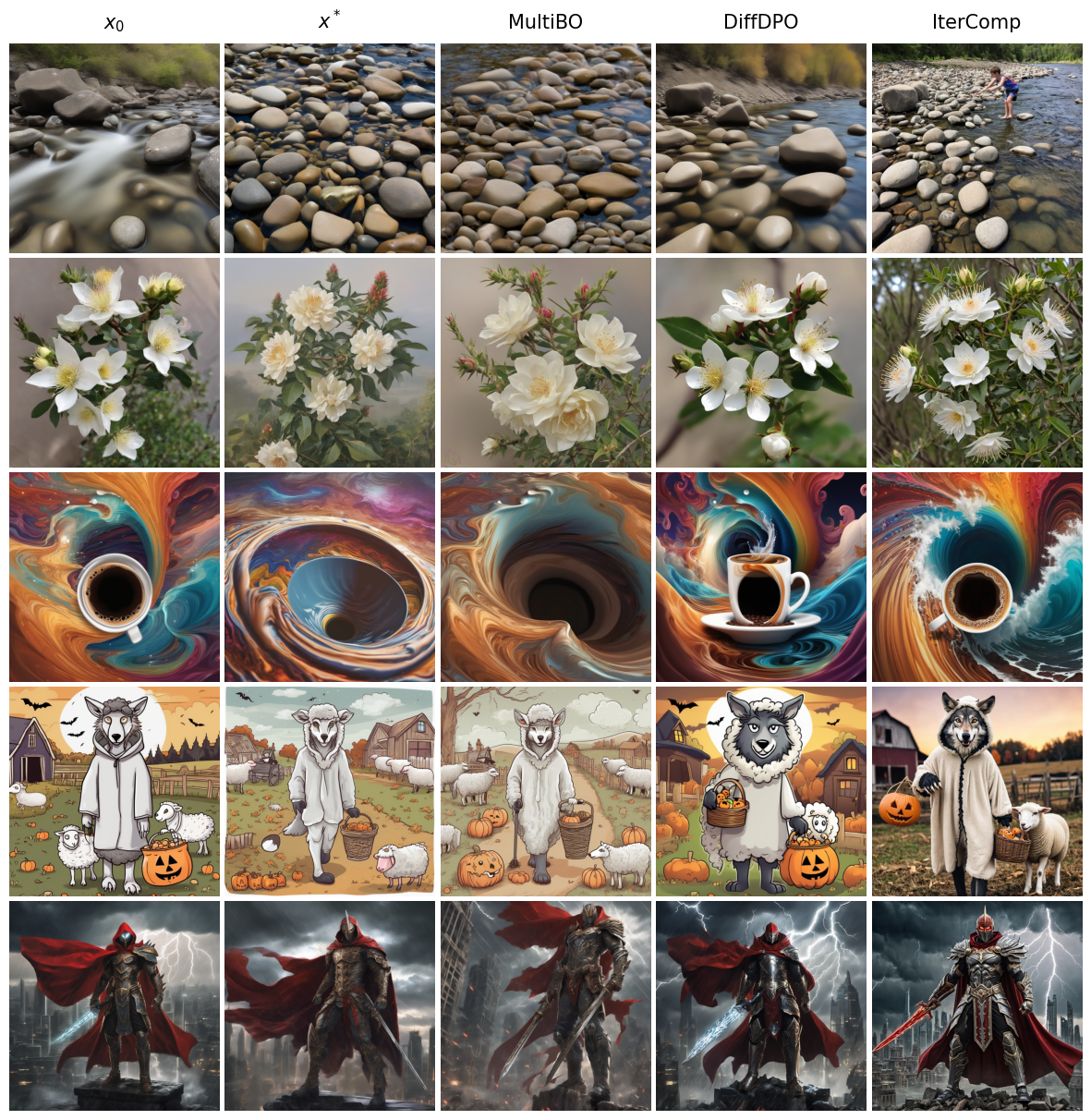}
    \caption{Qualitative results comparing {\name} ($B=50$), DiffusionDPO, and IterComp. For prompts: \textit{``The smooth, cool surface of the river rocks were perfect for skipping across the water's surface.", ``The fragrant flowers bloomed on the sturdy stem and the thorny bush.",``A swirling, multicolored portal emerges from the depths of an ocean of coffee, with waves of the rich liquid gently rippling outward. The portal engulfs a coffee cup, which serves as a gateway to a fantastical dimension. The surrounding digital art landscape reflects the colors of the portal, creating an alluring scene of endless possibilities.",``A wolf wearing a sheep halloween costume going trick-or-treating at the farm",``Amidst a stormy, apocalyptic skyline, a masked warrior stands resolute, adorned in intricate armor and a flowing cape. Lightning illuminates the dark clouds behind him, highlighting his steely determination. With a futuristic city in ruins at his back and a red sword in hand, he embodies the fusion of ancient valor and advanced technology, ready to face the chaos ahead."}}
    \label{fig:appendix_training}
\end{figure*}

\begin{figure*}
    \centering
    \includegraphics[width=1\linewidth]{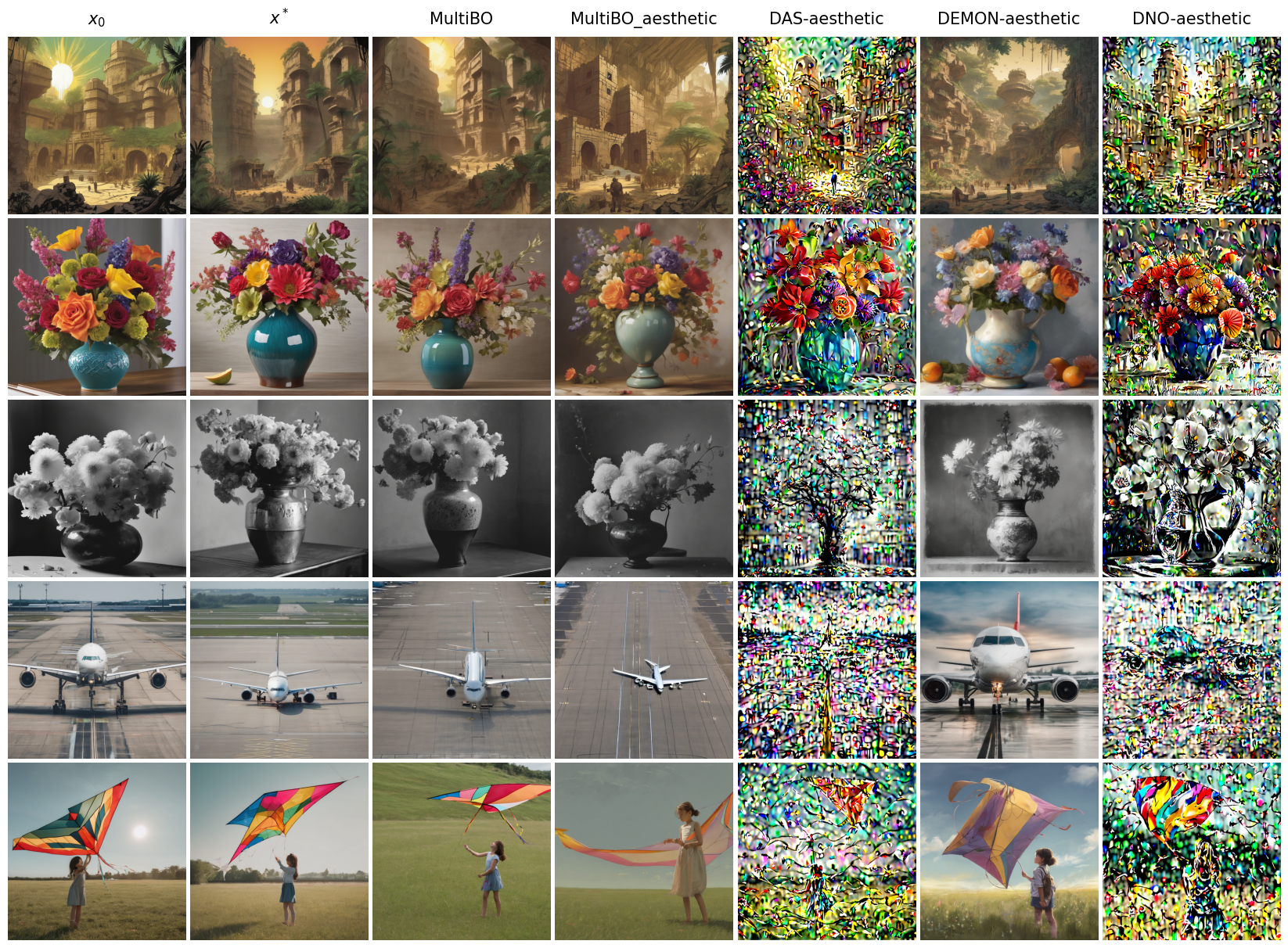}
    \caption{Qualitative results comparing {\name}  ($B=50$), {\name}$_{\text{Aesthetic}}$, {DAS}$_{\text{Aesthetic}}$, {DEMON}$_{\text{Aesthetic}}$, and DNO$_{\text{Aesthetic}}$. For prompts: \textit{``A comic book illustration by John Kirby depicting a jungle fortress surrounded by dirt walls in a marketplace setting with cinematic rays of sunlight.",``The smooth, glossy finish of the ceramic vase accentuated the vibrant colors of the flowers, a stunning centerpiece of beauty.",``A black and white photo of a steam of flowers inside a vase.",``An airplane on the runway of an airport.",``A girl is holding a large kite on a grassy field."}}
    \label{fig:appendix_aesthetic}
\end{figure*}

\begin{figure*}
    \centering
    \includegraphics[width=1\linewidth]{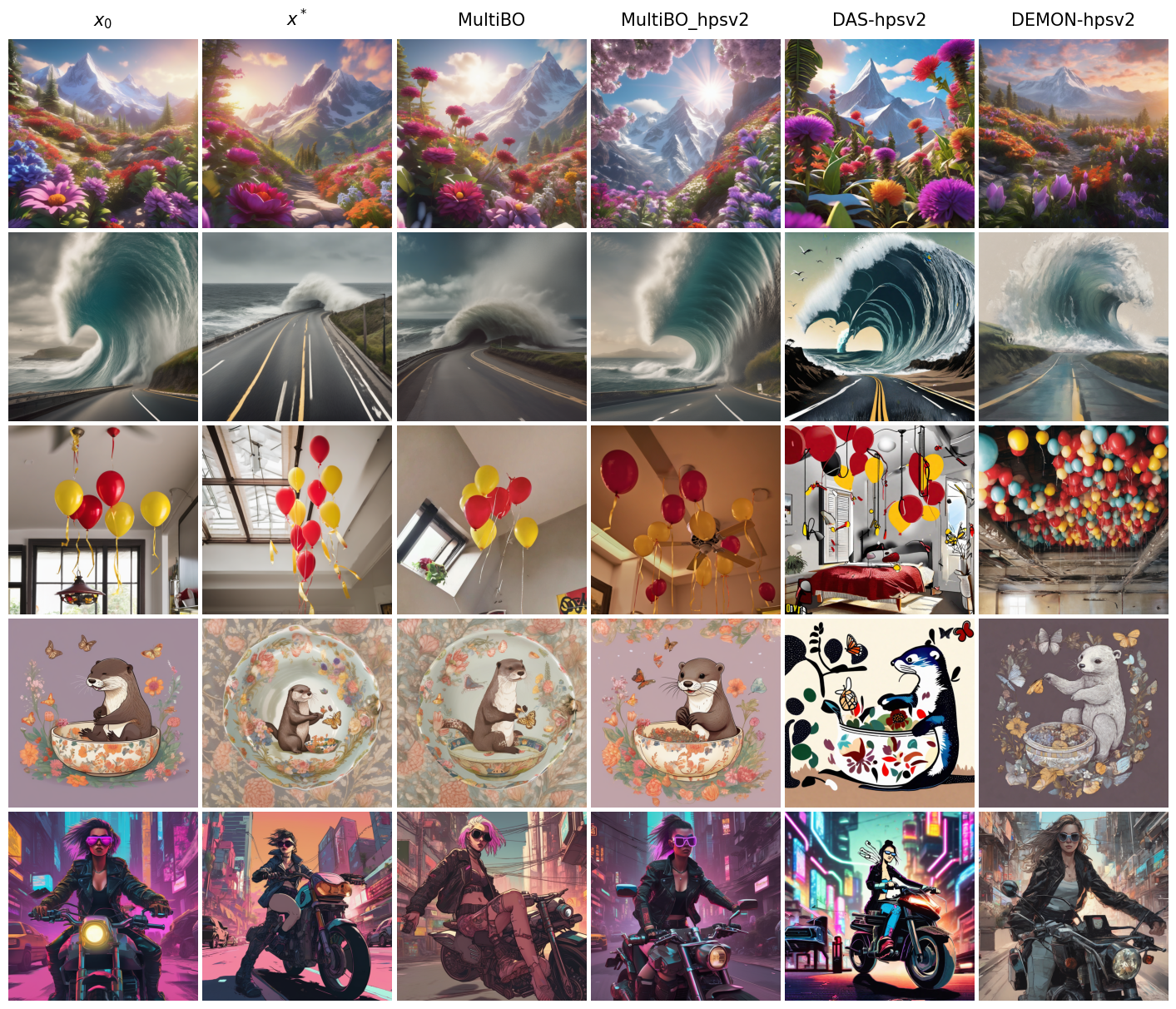}
    \caption{Qualitative results comparing {\name}  ($B=50$), {\name}$_{\text{HPSv2}}$, {DAS}$_{\text{HPSv2}}$, {DEMON}$_{\text{HPSv2}}$, and DNO$_{\text{HPSv2}}$. For prompts: \textit{``a stunning 3d render of towering, giant blooming plants with vibrant, colorful flowers on a picturesque mountain landscape.  sunlight dances on the petals, creating an enchanting scene as the wind gently sways the plants, with snow-capped peaks in the distance.",``a tidal wave approaching a coastal road",``red and yellow balloons hanging from a ceiling fan",``A flower patterned otter is playing with a butterfly shaped bowl",``A cyberpunk woman on a motorbike drives away down a street while wearing sunglasses."}}
    \label{fig:appendix_hpsv2}
\end{figure*}


\begin{figure*}
    \centering
    \includegraphics[width=1\linewidth]{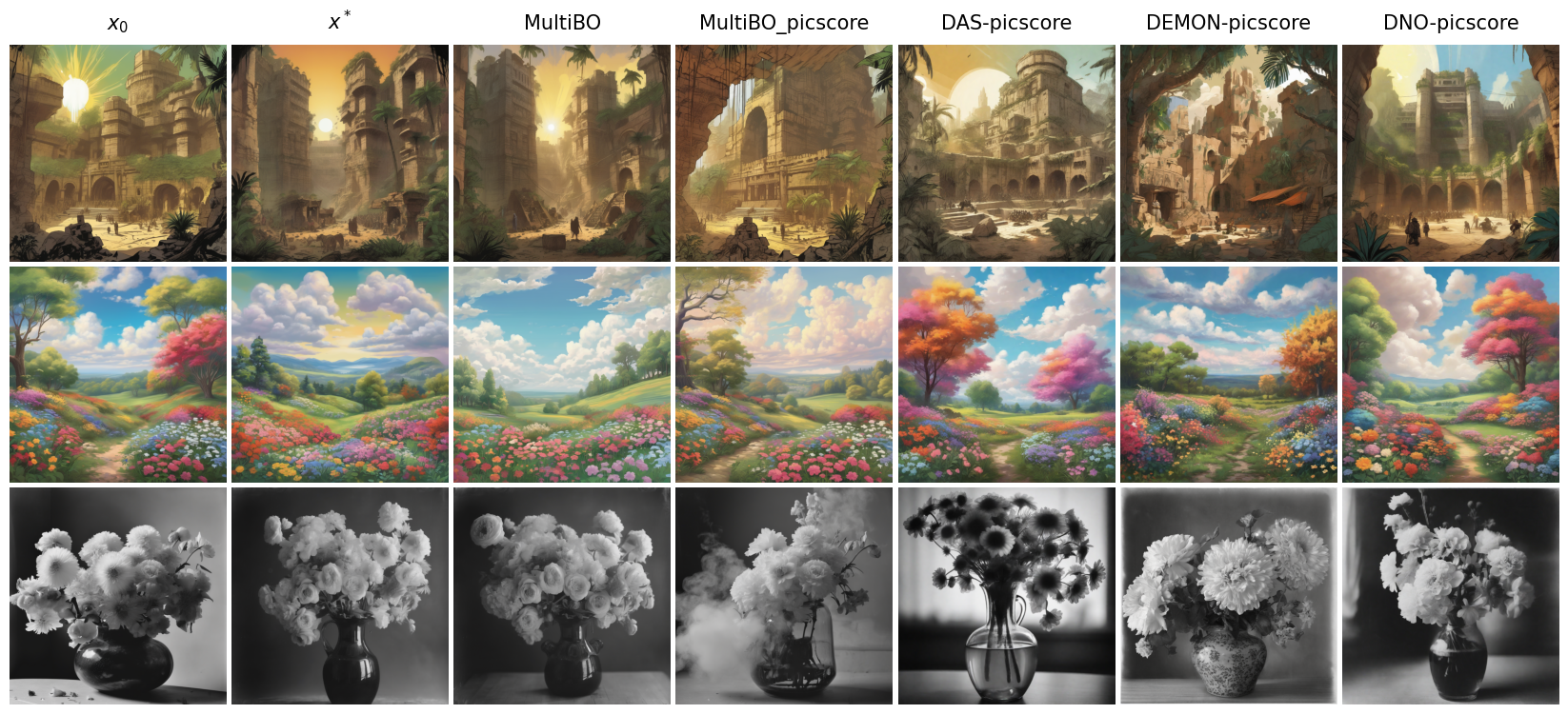}
    \caption{Qualitative results comparing {\name}  ($B=50$), {\name}$_{\text{PicScore}}$, {DAS}$_{\text{PicScore}}$, {DEMON}$_{\text{PicScore}}$, and DNO$_{\text{PicScore}}$. For prompts: \textit{``A comic book illustration by John Kirby depicting a jungle fortress surrounded by dirt walls in a marketplace setting with cinematic rays of sunlight.",``A peaceful, nature-filled landscape with vibrant flowers and trees and a serene cloud-filled sky.",``A black and white photo of a steam of flowers inside a vase."}}
    \label{fig:appendix_picscore}
\end{figure*}

\begin{figure*}
    \centering
    \includegraphics[width=1\linewidth]{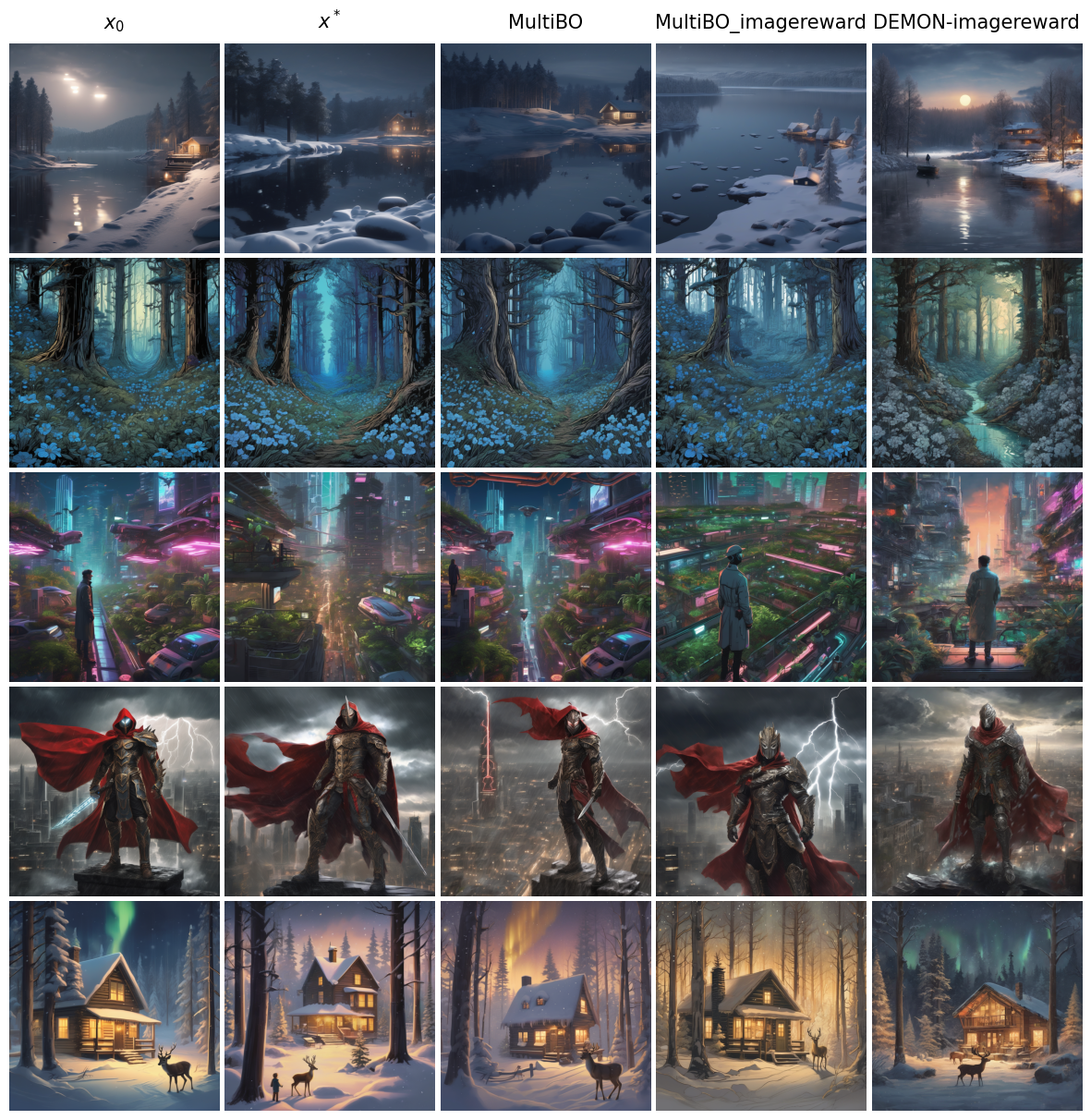}
    \caption{Qualitative results comparing {\name}  ($B=50$), {\name}$_{\text{ImageReward}}$, {DEMON}$_{\text{ImageReward}}$. For prompts: \textit{``A vividly realistic depiction of a snowy Swedish lake at night with hyper-detailed, cinematic-level artistry showcased on ArtStation.",``A forest with blue flowers illustrated in a digital matte style by Dan Mumford and M.W Kaluta.",``On the rooftop of a skyscraper in a bustling cyberpunk city, a figure in a trench coat and neon-lit visor stands amidst a garden of bio-luminescent plants, overlooking the maze of flying cars and towering holograms. Robotic birds flit among the foliage, digital billboards flash advertisements in the distance.",``Amidst a stormy, apocalyptic skyline, a masked warrior stands resolute, adorned in intricate armor and a flowing cape. Lightning illuminates the dark clouds behind him, highlighting his steely determination. With a futuristic city in ruins at his back and a red sword in hand, he embodies the fusion of ancient valor and advanced technology, ready to face the chaos ahead.",``A cozy winter cabin in a snowy forest at night. Warm yellow lights glow from the windows, and smoke gently rises from the chimney. A deer stands near the trees, watching as a child builds a snowman. In the sky, the northern lights shimmer above the treetops."}}
    \label{fig:appendix_imagereward}
\end{figure*}

\begin{figure*}
    \centering
    \includegraphics[width=1\linewidth]{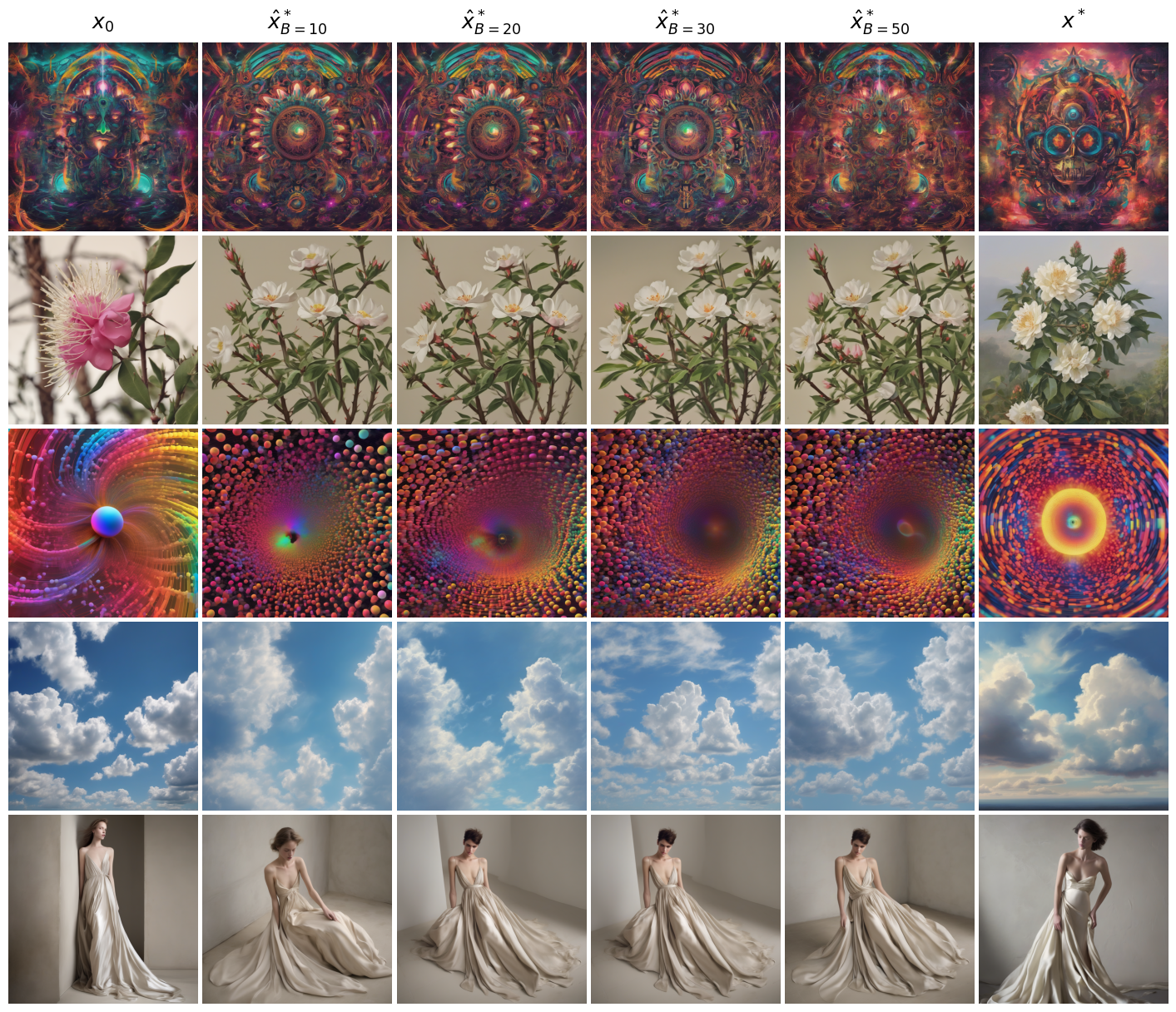}
    \caption{Qualitative Examples when {\name} falls short of reaching $x^*$ in $B=50$ iterations. For prompts: \textit{``Psytrance artwork featuring octane design.",``Beautiful flowering plant with big flowers.",``an electron cloud model is displayed in vibrant colors with a light spectrum background, showcasing the probability distribution of electrons around the nucleus. the image resembles digital art with pixelated elements, bringing a modern, educational twist to atomic structure visualization.",``The soft, fluffy clouds drifted lazily across the blue sky, a canvas of endless possibilities and imagination.",``The smooth silk gown flowed over the delicate skin and the rough floor."}}
    \label{fig:failure}
\end{figure*}